\PassOptionsToPackage{table}{xcolor}
\documentclass{sjtudenglab}
\usepackage{abstractoption}

\usepackage{amsmath,amsfonts,bm}

\def\eqref#1{equation~\ref{#1}}

\def\1{\bm{1}}

\DeclareMathAlphabet{\mathsfit}{\encodingdefault}{\sfdefault}{m}{sl}
\SetMathAlphabet{\mathsfit}{bold}{\encodingdefault}{\sfdefault}{bx}{n}

\usepackage{float}
\usepackage{needspace}
\usepackage{hyperref}
\hypersetup{pdfauthor={Haoqiang Kang, Yizhe Zhang, Nikki Lijing Kuang, Jiatao Gu, Yi-An Ma, Lianhui Qin},pdftitle={Uni-LaDiR: Latent Diffusion Unifies Multimodal Reasoning}}
\usepackage{url}
\usepackage[table]{xcolor}
\usepackage{booktabs}
\usepackage{multirow}
\usepackage{graphicx}
\usepackage{etoolbox}
\usepackage[normalem]{ulem}
\AtBeginEnvironment{table}{\setlength{\belowcaptionskip}{5pt}}
\AtBeginEnvironment{table*}{\setlength{\belowcaptionskip}{5pt}}
\newcommand{\method}{Uni-LaDiR}
\newcommand{\rev}[1]{#1}
\newcommand{\newrev}[1]{#1}
\newcommand{\rrev}[1]{#1}

\newcommand{\bluecell}[1]{#1}

\definecolor{tableheadbg}{HTML}{DFE7EF}
\definecolor{tableblue}{HTML}{E2ECF5}
\definecolor{tablepurple}{HTML}{F3EFE5}
\definecolor{tablegreen}{HTML}{EDF1EC}
\definecolor{tablegray}{HTML}{ECEEF0}
\definecolor{oursbg}{HTML}{F8F0DA}
\definecolor{oursaccent}{HTML}{182B49}

\title{Uni-LaDiR: Latent Diffusion Unifies\\Multimodal Reasoning}
\author[1]{Haoqiang Kang}
\author[2]{Yizhe Zhang}
\author[1]{Nikki Lijing Kuang}
\author[3]{Jiatao Gu}
\author[1]{Yi-An Ma}
\author[1]{Lianhui Qin}
\affiliation[1]{UC San Diego}
\affiliation[2]{Meta}
\affiliation[3]{University of Pennsylvania}
\displaytitle{{\color{optionaccent}Uni-LaDiR}: Latent Diffusion Unifies\\Multimodal Reasoning}
\displayauthors{\authorformat[1]{Haoqiang Kang},\enspace
\authorformat[2]{Yizhe Zhang},\enspace
\authorformat[1]{Nikki Lijing Kuang},\enspace
\authorformat[3]{Jiatao Gu},\enspace
\authorformat[1]{Yi-An Ma},\enspace
\authorformat[1]{Lianhui Qin}}
\displayaffiliations{\affiliationformat[1]{UC San Diego}\quad
\affiliationformat[2]{Meta}\quad
\affiliationformat[3]{University of Pennsylvania}}
\abstract{Multimodal models increasingly think with different modalities such as images, 3D point clouds, and robot states, not just text. Yet each modality is still encoded into its own representation space, creating a modality-switching gap whenever reasoning moves from one modality to another. In this paper, we introduce \method{} (\textbf{Uni}fied \textbf{La}tent \textbf{Di}ffusion \textbf{R}easoner), a framework that unifies different modalities into a shared latent space for multimodal reasoning. A unified encoder maps teacher reasoning steps from different modalities into latent thought tokens in a shared space, trained to extract the information needed for later reasoning steps and the final output. A diffusion reasoner, trained jointly with the encoder, generates these tokens at inference without teacher reasoning steps. Across \rev{eleven} vision-language model (VLM) benchmarks and two vision-language-action (VLA) suites, \method{} achieves relative gains over the strongest baselines of \textbf{7.3\%} on four mathematical and logical VLM benchmarks and \textbf{6.1\%} on RLBench manipulation tasks. Controlled comparisons show increasing gains as more teacher modalities are unified. These results suggest that unification improves multimodal reasoning by weaving it into a single thread, where the model predicts successive thoughts in a common representation space.}

\AtBeginDocument{\fancyfoot[C]{\thepage}}
\begin{document}
\raggedbottom
\renewcommand{\topfraction}{0.9}
\renewcommand{\bottomfraction}{0.85}
\renewcommand{\textfraction}{0.08}
\renewcommand{\floatpagefraction}{0.75}
\setlength{\textfloatsep}{14pt plus 2pt minus 2pt}
\setlength{\floatsep}{12pt plus 2pt minus 1pt}
\setlength{\intextsep}{12pt plus 2pt minus 1pt}
\setlength{\abovecaptionskip}{6pt}
\setlength{\bibsep}{2pt}
\makeatletter
\setlength{\@fptop}{0pt}
\setlength{\@fpsep}{12pt}
\makeatother
\maketitle

\begin{figure}[t]
  \centering
  \includegraphics[width=\textwidth]{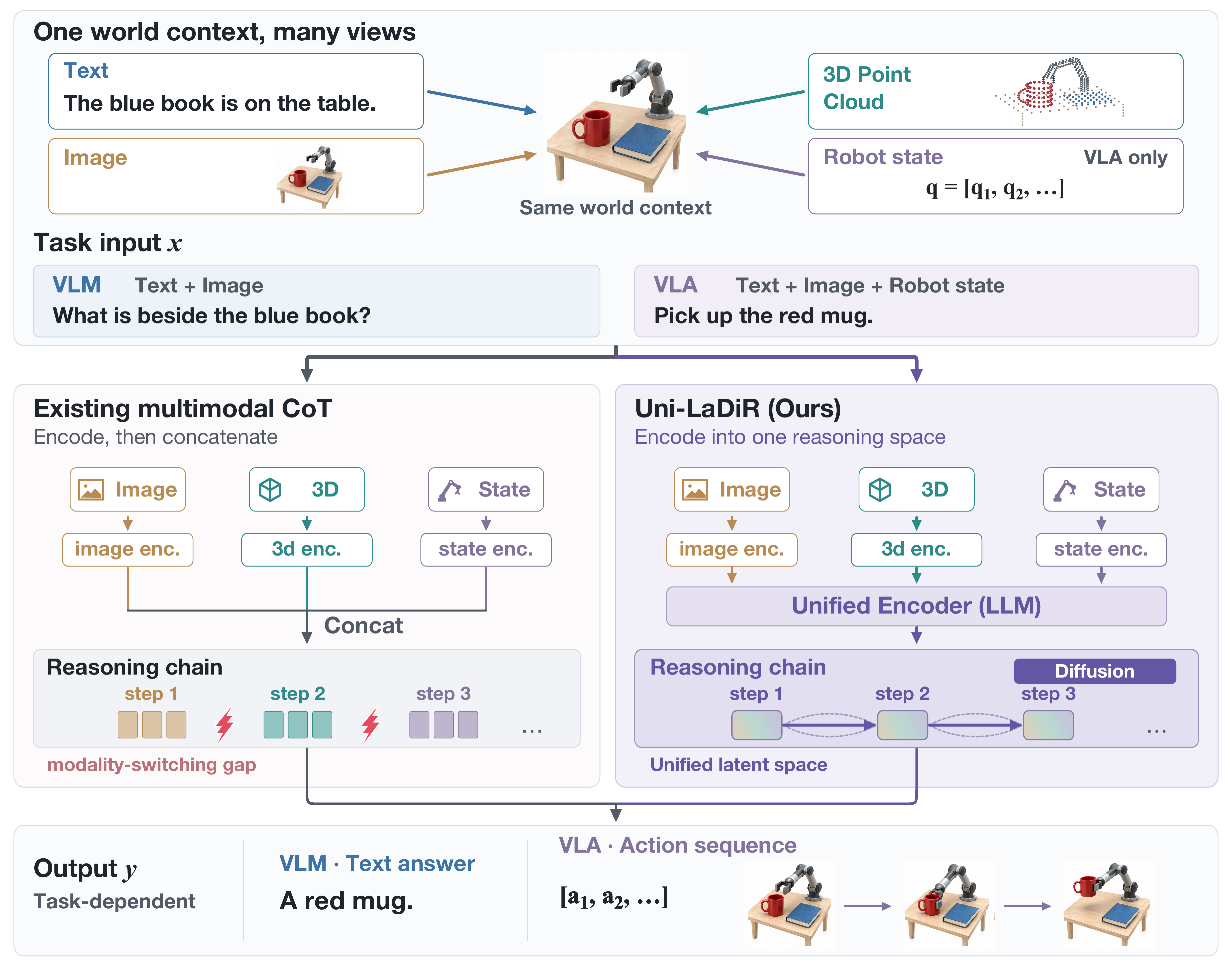}
  \caption{\textbf{Multimodal reasoning in a shared latent space.} Modality-specific reasoning (left) switches representations between steps. Uni-LaDiR (right) uses a shared latent space and diffusion to generate thought tokens for final outputs.}
  \label{fig:teaser}
\end{figure}

\begingroup
\color{black}
\section{Introduction}
\label{sec:introduction}
\providecommand{\introrevision}[1]{#1}

Multimodal reasoning often requires combining information from different modalities. An image captures visual appearance, language describes events and relations, 3D point clouds provide geometric structure, and robot states describe interaction with the environment. These modalities provide complementary information, and many tasks require using several of them together rather than relying on any single modality~\citep{lu2022scienceqa,lu2024mathvista,hao2025emma}. A capable multimodal model should therefore be able to integrate different modalities throughout its reasoning process~\citep{gao2024icot,gu2025thinkmorph}.

Recent work has explored several ways to incorporate multiple modalities into Chain-of-Thought (CoT) reasoning~\citep{wei2022cot}. In vision-language models (VLMs), reasoning chains can interleave text with visual steps, represented as either generated images~\citep{li2025mvot,gu2025thinkmorph} or latent visual tokens~\citep{li2025lvr,yang2025mirage}. Vision-language-action models (VLAs) similarly combine language, visual observations, and actions~\citep{brohan2023rt2,kim2024openvla,zhao2025cotvla}, with some methods additionally incorporating 3D information and robot states~\citep{zhen2024threedvla,qu2025spatialvla,liu2026last0}.

However, placing different modalities in the same reasoning sequence does not make their representations shared. Most existing methods still encode each modality in its own representation space and then interleave the resulting tokens in the LLM sequence~\citep{gao2024icot,li2025lvr,yang2025mirage,liu2026last0}. An image reasoning step may therefore be represented by visual features, while a 3D step or robot-state step uses a different representation. When reasoning moves from one modality to another, the model predicts the next step across different modality-specific representation spaces.

This motivates a simple idea: represent all intermediate reasoning steps in a shared latent space. The reasoning content may change across modalities, while its latent representation remains in the same space. A shared latent interface allows the reasoning model to predict successive steps in a common representation space. Later steps can use earlier latent representations through this shared interface, regardless of their source modality.

The main challenge is how to learn such a shared reasoning space. Adopting the representation of any single modality would retain its modality-specific structure. Instead, we learn the representation according to what later reasoning steps need. Based on this idea, we introduce \method{} (\textbf{Uni}fied \textbf{La}tent \textbf{Di}ffusion \textbf{R}easoner), a framework that maps reasoning steps from different modalities into a shared latent space. A unified encoder converts each teacher reasoning step into a fixed-sized block of latent thought tokens. Rather than reconstructing the original modality, these tokens are trained to predict subsequent reasoning steps and the final output. The learned representation is thus shaped by the information needed for future reasoning, rather than by the format of the original modality.

At inference time, the teacher reasoning steps are unavailable, so the model must generate these thought tokens from the task input and its previous reasoning. We use a diffusion reasoner to model this generation process~\citep{ho2020ddpm,rombach2022ldm,chi2023diffusionpolicy,kang2025ladir,kang2026ladi}. Diffusion is a natural choice here, since the same context may admit multiple valid next thoughts. We jointly train the unified encoder and diffusion reasoner so that the latent thoughts satisfy two requirements: they retain information useful for subsequent reasoning and the final output, while remaining predictable from the context available at inference time.

We evaluate \method{} on \rev{eleven} VLM reasoning benchmarks and two VLA manipulation suites. Compared with the strongest modality-specific latent reasoning baseline in each setting, \method{} improves mean accuracy by \textbf{7.3\%} relative across four mathematical and logical VLM benchmarks and mean success by \textbf{6.1\%} relative across ten RLBench manipulation tasks. Controlled experiments further isolate the effect of unification. Bringing more teacher modalities into the shared latent space consistently improves performance.
 These results support the central hypothesis of \method{}: multimodal reasoning benefits from keeping intermediate reasoning steps in a shared representation space, even when those steps originate from different modalities.

\par
\endgroup
\section{Preliminaries}
\label{sec:preliminaries}

\textcolor{black}{Existing multimodal reasoning methods often incorporate different modalities into their reasoning chains to combine complementary evidence for answering questions or planning actions~\citep{gao2024icot,li2025mvot,gu2025thinkmorph,zhao2025cotvla}. In latent multimodal CoT, intermediate content is represented by modality-specific latent blocks~\citep{li2025lvr,yang2025mirage,liu2026last0}. For a chain of $K$ steps, let $m_i$ denote the modality of step $i$ and $z_i^{(m_i)}\in\mathcal Z_{m_i}$ its latent block, where $\mathcal Z_m$ is the latent space of modality $m$. The blocks are concatenated into a single reasoning sequence:}
\begin{equation}
\textcolor{black}{[x;\,z_1^{(m_1)};\,\cdots;\,z_K^{(m_K)}]\;\longrightarrow\;y.}
\label{eq:latent-cot}
\end{equation}
\textcolor{black}{Here $x$ is the task input, $y$ is the task output, and $[\,;\,]$ denotes sequence concatenation. During training, modality-specific encoders provide teacher features $s_i$ to supervise the latent blocks; at inference, the model generates these blocks from $x$ and preceding blocks. The same formulation covers VLMs and VLAs (Figure~\ref{fig:teaser}): for VLMs, $x$ contains an image and a question, and $y$ is a text answer; for VLAs, $x$ contains visual observations (including point clouds when available), a language instruction, and the current robot state, and $y$ is an action sequence.} Because each block $z_i^{(m_i)}$ lies in the space of its own modality, predicting the next step can span different latent spaces, $\mathcal Z_{m_i}\to\mathcal Z_{m_{i+1}}$.

\section{Unified Latent Diffusion Reasoning}
\label{sec:method}

\begin{figure}[!htbp]
\centering
\includegraphics[width=\textwidth]{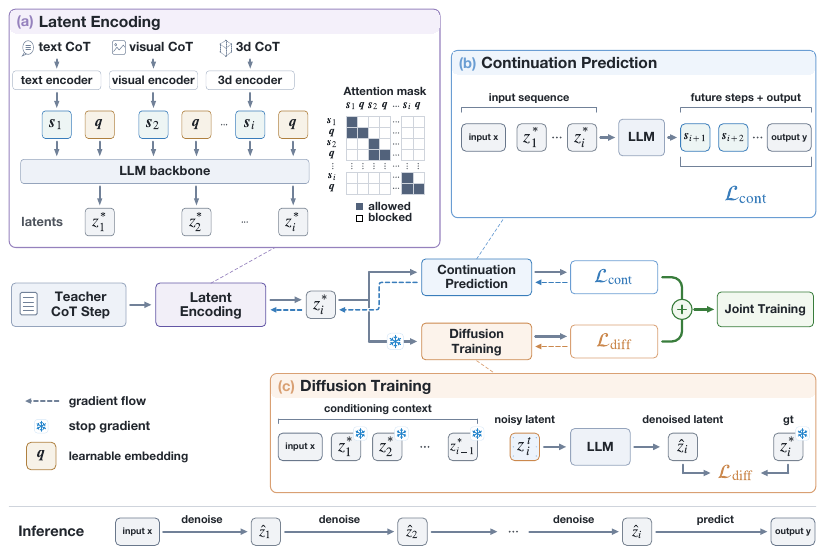}
\caption{\textbf{Uni-LaDiR overview.} During training, a shared backbone \textbf{(a)} encodes teacher steps into thought tokens, \textbf{(b)} \textcolor{black}{uses them to predict later teacher steps and the final output}, and \textbf{(c)} learns their generation by diffusion. \textcolor{black}{At inference, it generates thought blocks sequentially by diffusion, conditioned on the input and preceding blocks, then predicts the answer or action.}}
\label{fig:method}
\end{figure}

Uni-LaDiR encodes reasoning steps from all modalities into a shared latent space, allowing the model to predict successive steps through a common latent interface. In this section, we first formalize the requirements of this shared latent space (Section~\ref{sec:motivation}), then describe how teacher steps are mapped to shared thought tokens (Section~\ref{sec:construct}), how \textcolor{black}{continuation prediction trains} these tokens for subsequent reasoning and task prediction (Section~\ref{sec:consume}), and how latent diffusion learns to generate them from the context available at inference (Section~\ref{sec:diffusion}). Figure~\ref{fig:method} provides an overview.

\subsection{Motivation}
\label{sec:motivation}

\textcolor{black}{Recent work concatenates modality-specific reasoning steps into a single sequence. When adjacent steps use different modalities, their latent spaces differ, $\mathcal Z_{m_i}\ne\mathcal Z_{m_{i+1}}$, so the model predicts successive reasoning steps across modality-specific representation spaces (Eq.~\ref{eq:shared-transition}, top). We use a unified encoder $E_\theta$, with parameters $\theta$, to map teacher features $s_i$ to latent thought tokens $z_i^\star=E_\theta(s_i)$ in a shared latent space $\mathcal Z$:}
\begin{equation}
\textcolor{black}{\begin{array}{l@{\qquad}c@{\qquad}c}
& \text{reasoning} & \text{representation spaces} \\[3pt]
\text{modality-specific concat:}
& z_i^{(m_i)}\longrightarrow z_{i+1}^{(m_{i+1})}
& \underbrace{\mathcal Z_{m_i}\longrightarrow\mathcal Z_{m_{i+1}}}_{\text{switch}} \\[8pt]
\text{Uni-LaDiR:}
& z_i^\star\longrightarrow z_{i+1}^\star
& \underbrace{\mathcal Z\longrightarrow\mathcal Z}_{\text{no switch}}
\end{array}}
\label{eq:shared-transition}
\end{equation}
\textcolor{black}{Now both $z_i^\star$ and $z_{i+1}^\star$ lie in $\mathcal Z$ (Eq.~\ref{eq:shared-transition}, bottom): the thought changes, but its format stays the same. The model predicts successive steps within this shared space, using earlier latent tokens as context. Rather than adopting the features of any one modality, we learn this space from the reasoning itself. We train these thought tokens through \emph{continuation prediction} to retain information needed for later reasoning and the final output (\emph{reasoning sufficiency}), and jointly train a diffusion reasoner to generate them from the available context (\emph{context predictability}).}

\subsection{Encoding reasoning steps into a shared latent space}
\label{sec:construct}

We train a unified encoder to map each teacher reasoning step to the corresponding thought tokens $z_i^\star$ in the shared latent space (Figure~\ref{fig:method}a). After modality-specific encoding, we append a fixed-length block of learnable embeddings $q$ to each teacher feature $s_i$. Then our unified encoder maps the last hidden states at the $q$ positions to the clean thought tokens $z_i^\star$.

Every modality uses the same $q$, output projection, and number of thought tokens per block. During training, we pack multiple teacher steps and their corresponding copies of $q$ into one forward pass. The latent-encoding mask isolates the steps: each copy of $q$ attends only to itself and its associated teacher features, preventing information exchange across reasoning steps (Figure~\ref{fig:method}a). Thus, all teacher modalities are now encoded into a shared latent space, where later reasoning steps, rather than the source format, determine which information is retained (Section~\ref{sec:consume}).

\subsection{\textcolor{black}{Learning thought tokens through continuation prediction}}
\label{sec:consume}

\textcolor{black}{Continuation prediction} trains each thought block to preserve the information later reasoning will need. Given $x$ and the prefix $z_{<i}^\star=[z_1^\star;\ldots;z_{i-1}^\star]$, the backbone predicts the next teacher target $s_i^{\mathrm{GT}}$; given the complete sequence $z_{1:K}^\star=[z_1^\star;\ldots;z_K^\star]$, it predicts the final output $y$ (Figure~\ref{fig:method}b). Here $s_i$ denotes the teacher features mapped to a thought block, whereas $s_i^{\mathrm{GT}}$ denotes the fixed supervision target for that step: text tokens or continuous teacher features. A \textcolor{black}{continuation-prediction} attention mask computes these predictions in one forward pass while exposing each prediction only to its permitted thought-token prefix and hiding later blocks and raw teacher steps (Appendix~\ref{app:attention-masks}).

Let $\hat s_i$ and $\hat y$ denote the corresponding predictions. We optimize
\begin{equation}
\textcolor{black}{\mathcal{L}_{\mathrm{cont}}}
= \underbrace{\sum_{i=2}^{K}\ell(\hat s_i,s_i^{\mathrm{GT}})}_{\text{reasoning steps}}
+ \lambda_y\underbrace{\mathcal L_{\mathrm{out}}}_{\text{final output}}, \qquad
\ell(\hat u,u)
= \begin{cases}
\operatorname{CE}(\hat u,u), & u\text{ is text}, \\
\|\hat u-u\|_2^2, & \textcolor{black}{u\text{ is continuous}}.
\end{cases}
\label{eq:grounding}
\end{equation}
Here $u$ is a reasoning-step target, $\hat u$ its prediction, and $\operatorname{CE}$ the cross-entropy loss. The weight $\lambda_y$ scales the final-output loss $\mathcal L_{\mathrm{out}}$. \textcolor{black}{The continuous reasoning-step targets are image and 3D point-cloud features and robot-state vectors, supervised by the L2 term above. For VLMs, the final output $y$ is a text answer, so we use cross-entropy loss, $\mathcal L_{\mathrm{out}}=\operatorname{CE}(\hat y,y)$. For VLAs, $y$ is a continuous action sequence, which we model with a flow-matching action head; $\mathcal L_{\mathrm{out}}$ is its flow-matching loss.} The targets $s_i^{\mathrm{GT}}$ and $y$ are fixed, and any encoder used to construct target features is frozen. Gradients from $\mathcal{L}_{\mathrm{cont}}$ pass through $z_i^\star$ into the unified encoder and shared backbone, so the latent targets retain information useful for the continuation.

\subsection{Generating thought tokens with joint diffusion training}
\label{sec:diffusion}

The diffusion objective trains the backbone to generate each thought block from the context available at inference (Figure~\ref{fig:method}c). Diffusion models a conditional distribution over the next thought tokens, allowing different valid thoughts under the same input and preceding thought tokens~\citep{yu2025flow,kang2025ladir,kang2026ladi}. During training, we independently sample Gaussian noise $\epsilon_i\sim\mathcal N(0,I)$ for each block and time $t\sim\mathcal U[0,1]$, where $I$ is the identity covariance. Let $\operatorname{sg}(\cdot)$ denote stop-gradient, shown by snowflake markers in Figure~\ref{fig:method}c. We form the noisy block $z_i^t=(1-t)\epsilon_i+t\operatorname{sg}(z_i^\star)$, with target velocity $v_i^\star=z_i^\star-\epsilon_i$. Conditioned on the noisy block $z_i^t$, $x$, the detached prefix $\operatorname{sg}(z_{<i}^\star)$, and $t$, the shared backbone predicts the velocity $\hat v_i$.

We again process all reasoning steps in one forward pass. The diffusion mask restricts each noisy block to its own tokens, $x$, and the preceding clean thought blocks, preventing access to future thought blocks (Appendix~\ref{app:attention-masks}). We optimize the flow-matching objective~\citep{lipman2023flow}
\begin{equation}
\mathcal{L}_{\mathrm{diff}}
= \mathbb{E}_{t,\epsilon}\!\left[\sum_{i=1}^{K}
\|\hat v_i-\operatorname{sg}(v_i^\star)\|_2^2\right],
\label{eq:diffusion}
\end{equation}
The expectation is over $t$ and the block noises $\epsilon=(\epsilon_1,\ldots,\epsilon_K)$. Integrating the predicted velocity from $t=0$ to $1$ generates a thought block, denoted $z_i$. Thus, $t$ indexes denoising within a block, while $i$ indexes reasoning steps.

\paragraph{Joint training.} A useful thought token needs to retain task-relevant information and be predictable from the context available at inference. To meet both requirements, we share the LLM weights between the unified encoder and diffusion reasoner~\citep{duggal2026unite} and optimize
\begin{equation}
\mathcal{L}=\mathcal{L}_{\mathrm{cont}}
+\lambda\mathcal{L}_{\mathrm{diff}},
\label{eq:full}
\end{equation}
where $\lambda$ balances the two losses. The continuation loss trains the encoder to retain information needed for later reasoning and the final output. The diffusion loss trains the shared backbone to predict these representations from context, so its updates also change how the encoder represents teacher steps. Although we stop diffusion gradients through the clean targets and conditioning prefix, both losses still update the shared LLM weights. Diffusion training therefore also updates the unified encoder, allowing task utility and context predictability to jointly shape the latent representations.
\paragraph{Inference.}
At inference, the teacher reasoning steps and unified encoding path are not required. Starting from $x$, the model generates $z_1,\ldots,z_K$ sequentially by denoising one latent block at a time, with each block conditioned on the previously generated blocks; $K$ is not fixed, and generation stops once the model stops predicting \texttt{<BOT>} (Appendix~\ref{app:latent-sampling}) (Figure~\ref{fig:method}, bottom). The complete latent sequence then conditions prediction of the final answer or action $y$.
\section{Experiments}
\label{sec:experiments}

Our experiments test whether a shared latent space improves both visual reasoning and robotic manipulation. Section~\ref{sec:experimental-setup} introduces the benchmarks and baselines; Section~\ref{sec:main-results} reports task performance and manipulation inference efficiency; Section~\ref{sec:ablations} isolates the effect of unifying teacher modalities; and Section~\ref{sec:mechanism-ablations} ablates joint training, the encoder, and training objectives. \textcolor{black}{See Appendix~\ref{app:additional-experiments} for additional experimental results.}

\subsection{Evaluation protocol}
\label{sec:experimental-setup}

\paragraph{Visual reasoning.}
We train on Zebra-CoT~\citep{li2025zebracot}, which interleaves text and visual reasoning steps, and use Qwen2.5-VL-7B~\citep{bai2025qwen25vl} for the main experiment. We report accuracy on \rev{seven} vision-centric benchmarks: VisualPuzzles~\citep{song2025visualpuzzles}, ChartQA~\citep{masry2022chartqa}, V$^*$~\citep{wu2024vstar}, \rev{BLINK-Jigsaw~\citep{fu2024blink}}, MMVP~\citep{tong2024mmvp}, SAT~\citep{ray2024sat}, and CV-Bench~\citep{tong2024cambrian}. We also evaluate four mathematical or logical benchmarks: MathVista~\citep{lu2024mathvista}, MathVision~\citep{wang2024mathvision}, VisuLogic~\citep{xu2025visulogic}, and EMMA~\citep{hao2025emma}.

\paragraph{VLA manipulation.}
We evaluate LIBERO~\citep{liu2024libero} and RLBench~\citep{james2020rlbench}. In this setting, future images, 3D point clouds (RLBench only), and robot states serve as training-only teacher CoTs; continuous actions remain task outputs. Following \cite{liu2026last0}, our VLA backbone is a mixture-of-transformers (MoT) that couples two Janus-Pro-1.5B experts (3.3B parameters total), and actions are produced by a flow-matching action head trained with the same action objective. We report success rate and, on RLBench, control frequency without action chunking.

\paragraph{Baselines.}
For VLMs, we compare with the Qwen2.5-VL base model and explicit reasoners DeepEyes~\citep{zheng2025deepeyes}, PixelReasoner~\citep{su2025pixelreasoner}, Vision-R1~\citep{huang2025visionr1}, and Open-Vision-Reasoner~\citep{wei2025ovr}. Latent reasoning baselines include LVR~\citep{li2025lvr}, Mirage~\citep{yang2025mirage}, CoVT~\citep{qin2025covt}, Monet~\citep{wang2026monet}, \mbox{VaLR-M~\citep{jeon2026valr}}, SLVR~\citep{fan2026slvr}, Mull-Tokens~\citep{ray2025mulltokens}, and ILVR~\citep{viveiros2026ilvr}.
For VLAs, we consider three groups. \emph{Action policies} include OpenVLA~\citep{kim2024openvla}, SpatialVLA~\citep{qu2025spatialvla}, CogACT~\citep{li2024cogact}, $\pi_{0.5}$~\citep{pi05}, OpenVLA-OFT~\citep{kim2025openvlaoft}, HybridVLA~\citep{liu2025hybridvla}, and UniVLA~\citep{wang2025univla}. \emph{Explicit reasoning models} include ManipLLM~\citep{li2023manipllm}, WorldVLA~\citep{cen2025worldvla}, CoT-VLA~\citep{zhao2025cotvla}, and FlowVLA~\citep{zhong2025flowvla}. \emph{Latent reasoning models} include FiS-VLA~\citep{chen2025fisvla}, \textcolor{black}{FUTURE-VLA~\citep{xu2026futurevla}}, ConsisVLA-4D~\citep{li2026consisvla}, LaST$_0$~\citep{liu2026last0}, AVA-VLA~\citep{lei2026avavla}, and PearlVLA with and without PRL~\citep{yang2026pearlvla}. Full settings appear in Appendix~\ref{app:implementation}.

\begin{table}[!h]
  \centering
  \caption{\textbf{Comparison of \method{} and baselines on RLBench.} We report success rates (\%) and inference speed (Hz). Best results are highlighted in bold.}
  \label{tab:rlbench}
  \footnotesize
  \setlength{\tabcolsep}{2.5pt}
  \resizebox{\textwidth}{!}{%
  \begin{tabular}{lc*{10}{r}!{\tablesummarysep}>{\color{optiontitleink}}r@{\hspace{6pt}}rr}
  \toprule
  \textbf{\textcolor{optiontitleink}{Method}} & \textbf{Size} & \textbf{Box} & \textbf{Laptop} & \textbf{Toilet} & \textbf{Sweep} & \textbf{Fridge} & \textbf{Phone} & \textbf{Umbrella} & \textbf{Frame} & \textbf{Wine} & \textbf{Plants} & \textbf{Mean} & \textbf{\rrev{Std.}} & \textbf{Hz} \\
  \midrule
  \rowcolor{tableblue}\multicolumn{15}{c}{\tablegroup{Direct Policies}} \\
  OpenVLA & 7B & 60.0 & 35.0 & 75.0 & 55.0 & 85.0 & 20.0 & 30.0 & 15.0 & 20.0 & 5.0 & 40.0 & \rrev{2.0} & 6.3 \\
  SpatialVLA & -- & 80.0 & 70.0 & 85.0 & 20.0 & 80.0 & 15.0 & 25.0 & 40.0 & 15.0 & 30.0 & 46.0 & \rrev{3.0} & 7.9 \\
  CogACT & 7B & 90.0 & 80.0 & 95.0 & 50.0 & 85.0 & 50.0 & 55.0 & 45.0 & 30.0 & 25.0 & 61.0 & \rrev{4.0} & 9.8 \\
  $\pi_{0.5}$ & 3B & 90.0 & 95.0 & 85.0 & 75.0 & \textbf{100.0} & 5.0 & 10.0 & \textbf{80.0} & 75.0 & 35.0 & 65.0 & \rrev{4.0} & 13.8 \\
  HybridVLA & 7B & 85.0 & 95.0 & \textbf{100.0} & \textbf{90.0} & \textbf{100.0} & 50.0 & 50.0 & 70.0 & 50.0 & 50.0 & 74.0 & \rrev{4.0} & 6.1 \\
  
  \midrule
  \rowcolor{tableblue}\multicolumn{15}{c}{\tablegroup{Explicit Reasoning}} \\
  ManipLLM & 7B & 50.0 & 80.0 & 40.0 & 20.0 & 80.0 & 35.0 & 10.0 & 25.0 & 15.0 & 20.0 & 38.0 & \rrev{4.0} & 2.2 \\
  CoT-VLA & -- & 95.0 & 75.0 & \textbf{100.0} & 80.0 & 65.0 & 50.0 & 40.0 & 50.0 & 55.0 & 50.0 & 66.0 & \rrev{3.0} & 1.1 \\
  
  \midrule
  \rowcolor{tableblue}\multicolumn{15}{c}{\tablegroup{Latent Reasoning}} \\
  FiS-VLA & 7B & \textbf{100.0} & \textbf{100.0} & 95.0 & 55.0 & 90.0 & 50.0 & 50.0 & 70.0 & 55.0 & 20.0 & 69.0 & \rrev{3.0} & \textbf{21.9} \\
  LaST$_0$ & 3.3B & 95.0 & 95.0 & \textbf{100.0} & 80.0 & 85.0 & 75.0 & 75.0 & 70.0 & 85.0 & 60.0 & 82.0 & \rrev{3.0} & 15.4 \\
  
  \midrule
  \textcolor{optiontitleink}{\textbf{Uni-LaDiR (Ours)}} & 3.3B & \textbf{100.0} & \textbf{100.0} & \textbf{100.0} & 85.0 & 90.0 & \textbf{80.0} & \textbf{80.0} & 75.0 & \textbf{90.0} & \textbf{70.0} & \textcolor{black}{\textbf{87.0}} & \rrev{3.0} & 16.5 \\
  \bottomrule
  \end{tabular}}
  \vspace{+2mm}
  \end{table}

\begin{table}[!b]
\centering
\caption{\textbf{Comparison on VLM reasoning benchmarks.} Accuracy (\%) of \method{} and baselines with Qwen2.5-VL-7B on \textbf{(a)} vision-centric and \textbf{(b)} mathematical and logical benchmarks. Best results are highlighted in bold.}
\label{tab:vlm-main}
\label{tab:visual-published}
\label{tab:visual-controlled}
\label{tab:visual-extended}
\footnotesize
\setlength{\tabcolsep}{3pt}
\renewcommand{\arraystretch}{1.10}

\textbf{(a) Vision-Centric Perception, Search, and Spatial Reasoning.}\par
\begin{tabular*}{\textwidth}{@{\extracolsep{\fill}}l*{7}{r}!{\tablesummarysep}>{\color{optiontitleink}}r@{}}
\toprule
\textcolor{optiontitleink}{Method} & VisPuz. & ChartQA & V$^*$ & BLINK-J & MMVP & SAT & CV-B & Avg. \\
\midrule
\rowcolor{tableblue}\multicolumn{9}{c}{\tablegroup{Explicit Text and Visual Reasoning}} \\
Qwen2.5-VL (base) & \bluecell{34.8} & 78.1 & 76.4 & 59.3 & 77.3 & 51.3 & 75.2 & \rev{64.6} \\
DeepEyes & 48.0 & 82.5 & \rrev{85.3} & 72.0 & 78.0 & 64.0 & 84.0 & \rrev{73.4} \\
PixelReasoner & 44.0 & 81.5 & \rrev{81.7} & 70.0 & 76.0 & 60.0 & 82.0 & \rrev{70.7} \\

\midrule
\rowcolor{tableblue}\multicolumn{9}{c}{\tablegroup{Latent Reasoning}} \\
LVR & 46.0 & 81.0 & \bluecell{81.7} & \bluecell{52.0} & \bluecell{71.7} & 60.0 & 84.0 & \rev{68.1} \\
Mirage & 44.0 & 80.5 & 83.8 & \rrev{68.0} & 74.0 & \bluecell{72.0} & 83.0 & \rrev{72.2} \\
CoVT & 41.0 & 79.5 & \bluecell{78.0} & 68.0 & \bluecell{58.7} & 61.3 & 80.0 & \rev{66.7} \\
Monet & 35.0 & 79.0 & 83.3 & 65.3 & \bluecell{50.0} & 55.3 & \bluecell{71.1} & \rev{62.7} \\
VaLR-M & 45.0 & 82.0 & \bluecell{86.9} & 72.0 & \bluecell{60.3} & 65.3 & 87.6 & \rev{71.3} \\
SLVR & \bluecell{34.2} & 77.2 & 82.2 & 68.0 & 74.0 & 62.0 & 83.0 & \rev{68.7} \\
Mull-Tokens & 48.0 & 81.0 & 83.8 & \bluecell{74.7} & 76.0 & \bluecell{\textbf{77.0}} & 84.0 & \rev{74.9} \\

\midrule
\textcolor{optiontitleink}{\textbf{Uni-LaDiR (Ours)}} & \textbf{57.4} & \textbf{84.2} & \rrev{\textbf{89.5}} & \textbf{75.3} & \textbf{82.0} & \rrev{72.0} & \rrev{\textbf{88.8}} & \rrev{\textbf{78.5}} \\
\bottomrule
\end{tabular*}

\vspace{2mm}

\textbf{(b) Visual Mathematics and Logic.}\par
\begin{tabular*}{\textwidth}{@{\extracolsep{\fill}}l*{4}{r}!{\tablesummarysep}>{\color{optiontitleink}}r@{}}
\toprule
\textcolor{optiontitleink}{Method} & MathVista & MathVision & VisuLogic & EMMA & Avg. \\
\midrule
\rowcolor{tableblue}\multicolumn{6}{c}{\tablegroup{Explicit Text and Visual Reasoning}} \\
Qwen2.5-VL (base) & 68.2 & \bluecell{25.4} & 26.0 & 26.0 & 36.4 \\
Vision-R1 & \textbf{73.5} & 41.0 & \bluecell{15.2} & \bluecell{24.3} & 38.5 \\
Open-Vision-Reasoner & 69.5 & 51.8 & 27.2 & 30.0 & 44.6 \\
DeepEyes & 70.1 & 26.6 & 27.8 & 29.0 & 38.4 \\
PixelReasoner & 68.9 & 34.2 & 23.4 & \bluecell{23.5} & 37.5 \\

\midrule
\rowcolor{tableblue}\multicolumn{6}{c}{\tablegroup{Latent Reasoning}} \\
Mirage & 68.6 & 40.5 & 26.6 & 27.0 & 40.7 \\
Monet & 62.5 & 33.8 & \bluecell{10.6} & \bluecell{22.1} & 32.3 \\
ILVR & 71.1 & 49.0 & 29.3 & 33.3 & 45.7 \\

\midrule
\textcolor{optiontitleink}{\textbf{Uni-LaDiR (Ours)}} & 73.2 & \textbf{54.3} & \textbf{33.0} & \textbf{35.6} & \textcolor{black}{\textbf{49.0}} \\
\bottomrule
\end{tabular*}
\vspace{+4mm}

\end{table}

\newpage
\subsection{Main results}
\label{sec:main-results}

\paragraph{VLM visual reasoning.}\looseness=-1
On the \rev{seven} vision-centric benchmarks, \method{} improves average accuracy over Qwen2.5-VL-7B by \rrev{21.4\%}; on the four mathematical and logical benchmarks, the relative gain is 34.7\% (Table~\ref{tab:vlm-main}). Reasoning through shared latent thoughts thus strengthens both perception-heavy and multi-step mathematical reasoning. \method{} also outperforms the strongest latent baseline in each group, by \rrev{4.7\%} over Mull-Tokens and 7.3\% over ILVR on average. Since these baselines tie each latent step to modality-specific targets, the gains suggest that predicting successive reasoning steps in a shared latent space is more effective than reasoning across modality-specific representations. The gains over modality-specific baselines reach 30.3\% on VisualPuzzles over Mirage, and 10.8\% on MathVision and 12.6\% on VisuLogic over ILVR. These tasks require combining visual evidence with multi-step abstract reasoning, where a shared latent space lets later steps build on earlier visual findings regardless of their source modality.

\vspace{+0mm}

\paragraph{VLA manipulation.}
\method{} improves average success over the strongest prior latent policies by 0.45 percentage points over PearlVLA + PRL on LIBERO and 6.1\% relative to LaST$_0$ on RLBench (Tables~\ref{tab:libero} and~\ref{tab:rlbench}). LaST$_0$ shares our backbone, demonstrations, and evaluation protocol but keeps a separate latent space for each modality (Appendix~\ref{app:protocols}), so the RLBench gain mainly reflects the benefit of a shared latent space. On RLBench, \method{} improves nine of ten tasks and matches the remaining one; the largest gain is 16.7\% on Plants. The benefit is therefore not tied to a few tasks, but holds broadly when reasoning spans images, 3D point clouds, and robot states. Without action chunking, \method{} runs at \textbf{16.5 Hz}, a 7.1\% higher inference rate than LaST$_0$ and \textbf{15.0$\times$} the rate of explicit CoT-VLA. Reasoning in a compact shared latent space thus improves success without sacrificing control frequency, unlike explicit CoT that decodes intermediate images.

\begin{table}[!htbp]
\centering
\caption{\textbf{Comparison of \method{} and baselines on LIBERO.} We report success rates (\%) across four suites. Best results are highlighted in bold.}
\label{tab:libero}
\scriptsize
\setlength{\tabcolsep}{5.2pt}
\renewcommand{\arraystretch}{1.10}
\begin{tabular*}{\textwidth}{@{\extracolsep{\fill}}lc*{4}{r}!{\tablesummarysep}>{\color{optiontitleink}}r@{}}
\toprule
\textbf{\textcolor{optiontitleink}{Method}} & \textbf{Size} & \textbf{Spatial} & \textbf{Object} & \textbf{Goal} & \textbf{Long} & \textbf{Mean} \\
\midrule
\rowcolor{tableblue}\multicolumn{7}{c}{\tablegroup{Direct Policies}} \\
UniVLA & -- & 92.6 & 93.8 & 86.6 & 63.0 & 84.0 \\
OpenVLA & 7B & 84.7 & 88.4 & 79.2 & 53.7 & 76.5 \\
SpatialVLA & -- & 88.2 & 89.9 & 78.6 & 55.5 & 78.1 \\
CogACT & 7B & 97.2 & 98.0 & 90.2 & 88.8 & 93.6 \\
$\pi_{0.5}$ & 3B & 98.8 & 98.2 & 98.0 & 92.4 & 96.9 \\
OpenVLA-OFT & 7B & 97.6 & 98.4 & 97.9 & 94.5 & 97.1 \\

\midrule
\rowcolor{tableblue}\multicolumn{7}{c}{\tablegroup{Explicit Reasoning}} \\
WorldVLA & -- & 85.6 & 89.0 & 82.6 & 59.0 & 79.1 \\
CoT-VLA & -- & 87.5 & 91.6 & 87.6 & 69.0 & 83.9 \\
FlowVLA & 8.5B & 93.2 & 95.0 & 91.6 & 72.6 & 88.1 \\

\midrule
\rowcolor{tableblue}\multicolumn{7}{c}{\tablegroup{Latent Reasoning}} \\
\textcolor{black}{FUTURE-VLA} & 4B & 89.6 & 99.0 & 95.2 & 81.2 & 91.3 \\
ConsisVLA-4D & 7B & 98.8 & \textbf{99.8} & 98.0 & 95.6 & 98.1 \\
LaST$_0$ & 3.3B & 99.2 & 99.6 & 98.0 & 95.6 & 98.1 \\
AVA-VLA & 7B & 97.8 & 99.4 & 97.8 & \textbf{98.1} & 98.3 \\
PearlVLA & 7B & 99.2 & 99.6 & 98.2 & 96.8 & 98.5 \\
PearlVLA + PRL & 7B & 99.4 & \textbf{99.8} & 98.4 & 97.2 & 98.7 \\

\midrule
\textcolor{optiontitleink}{\textbf{Uni-LaDiR (Ours)}} & 3.3B & \textbf{100.0} & 99.0 & \textbf{100.0} & 97.6 & \textcolor{black}{\textbf{99.2}} \\
\bottomrule
\end{tabular*}
\vspace{+2mm}
\end{table}

\Needspace{8\baselineskip}
\subsection{Analysis}
\label{sec:ablations}
\begin{figure}[!htbp]
  \centering

  \includegraphics[width=\textwidth]{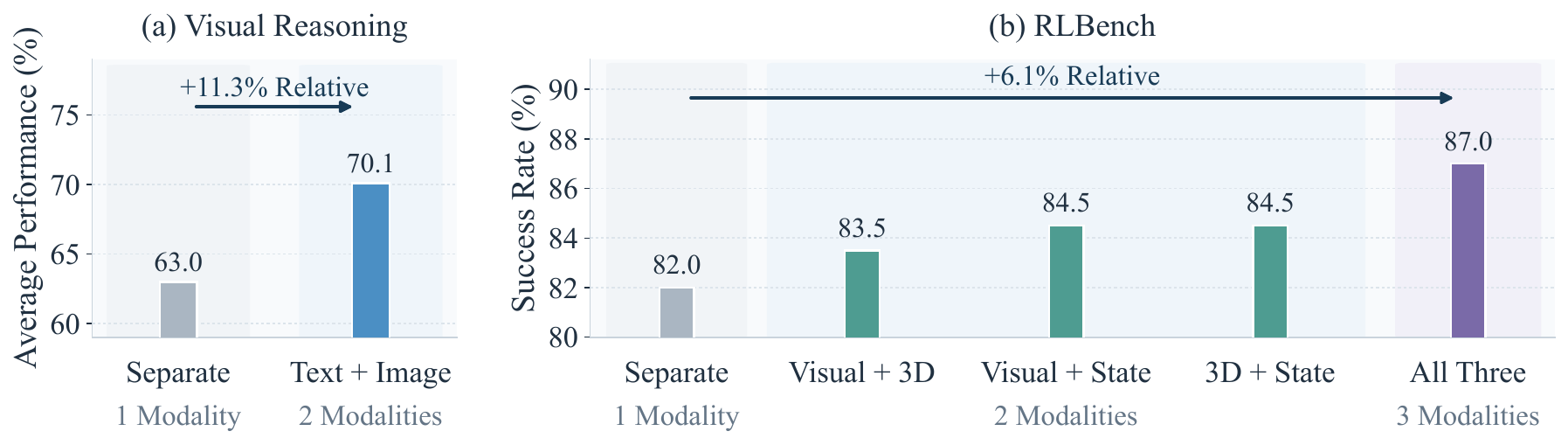}
  \caption{\textbf{Results of unifying more modalities.} \textbf{(a)} Qwen2.5-VL-7B mean accuracy on V$^*$, MMVP, MathVista, and EMMA. \textbf{(b)} RLBench success with different modality groupings.}
  \label{fig:ablation-sharing}
\end{figure}

\paragraph{Unifying more modalities.}
We first test whether all modalities need to be unified, or whether a
subset suffices.
The Separate baseline is the continuation-trained separate-encoding variant
in Figure~\ref{fig:ablation-objective}. We progressively route teacher modalities
through the unified encoder: partial variants unify selected modalities while the other modalities keep their own LLM encoders, and the full variant unifies all
modalities. Teacher modalities and continuation supervision remain fixed.
On the VLM benchmarks, unifying text and image improves mean accuracy by
\rrev{11.3\%} relative to separate encoders (Figure~\ref{fig:ablation-sharing}a).
On RLBench, unifying any pair of the visual, 3D, and state modalities yields
1.8--3.0\% relative gains, whereas unifying all three yields 6.1\% over separate encoders and 3.0\% over the best pair (Figure~\ref{fig:ablation-sharing}b).
These results show that gains grow as more modalities are unified, and that
leaving any one modality out limits the benefit.

\subsection{Ablation Study}

\label{sec:mechanism-ablations}

\begingroup
\paragraph{\textcolor{black}{Joint training}.}
To test whether the encoder and reasoner benefit from adapting together, we
compare separate-stage and joint training under the same unified architecture
and continuation supervision. Separate-stage training learns the encoder first
and then fixes it while training the reasoner; joint training updates both
together. Joint training improves the six-benchmark average by \rrev{12.7\%}
relative to separate stages (Figure~\ref{fig:ablation-training}), including
8.8\% on EMMA and 15.0\% on RLBench. These results show that
learning the representation and its predictor together outperforms fixing the
representation before training the reasoner.

\begin{figure}[!htbp]
  \centering
  \includegraphics[width=\textwidth]{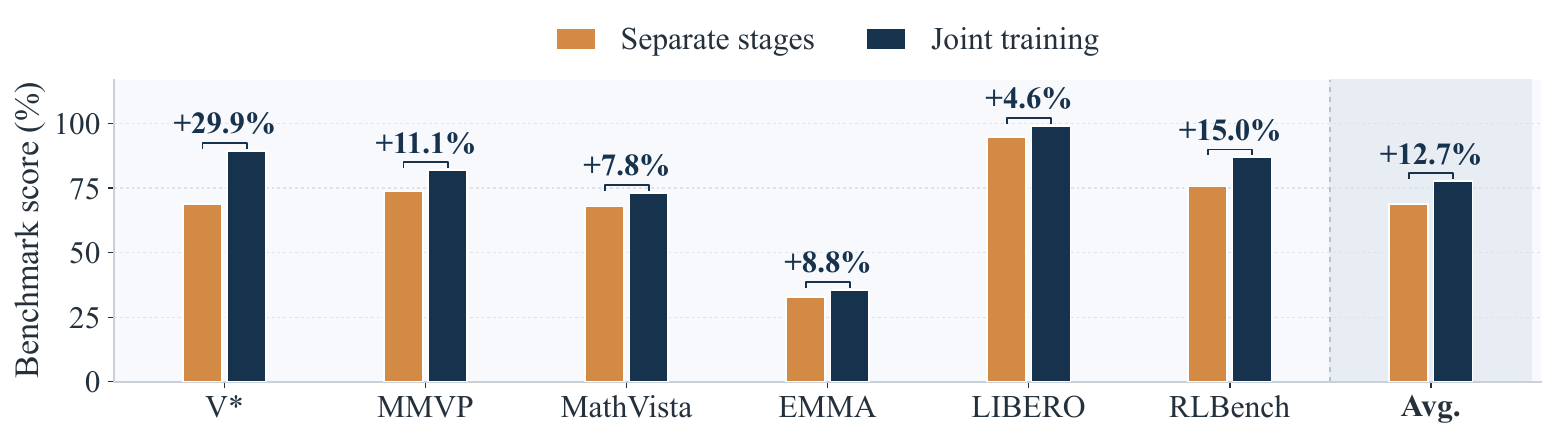}
  \caption{Comparison of separate-stage and joint training of LLM unified encoder and diffusion reasoner.}
  \label{fig:ablation-training}
\end{figure}
\endgroup

\begin{figure}[!htbp]
  \centering
  \includegraphics[width=\textwidth]{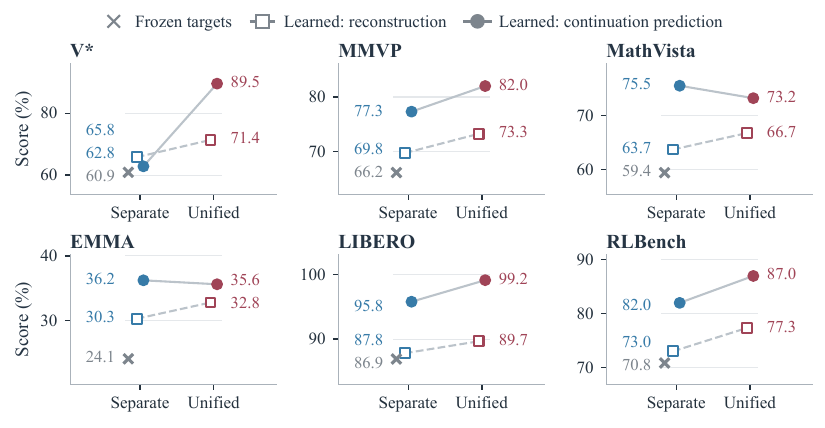}
  \caption{Comparison of separate and unified encoding with frozen targets or targets learned by reconstruction or continuation.}
  \label{fig:ablation-objective}
\end{figure}

\paragraph{Unified encoding.}
To test whether the unified encoder improves reasoning, we compare it with separate encoding under both reconstruction and continuation supervision. In separate encoding, each modality has its own LLM encoder with the same architecture, whereas unified encoding shares one LLM encoder across all modalities; all other settings are identical. Unified encoding improves the
six-benchmark average by 3.5 percentage points under reconstruction and 6.1
under continuation (Figure~\ref{fig:ablation-objective}). These results show that unified encoding improves the average under both losses.

\Needspace{6\baselineskip}
\paragraph{Latent target learning.}
To test whether learning the latent targets helps reasoning, we compare frozen
teacher targets with targets learned by reconstruction or continuation.
Reconstruction preserves source features, while continuation predicts later
reasoning steps; both learned variants keep the same final-output supervision.
With separate encoding, reconstruction and continuation improve the
six-benchmark average over frozen targets by 3.7 and 10.2 percentage points
(Figure~\ref{fig:ablation-objective}), and continuation outperforms
reconstruction by 6.6 points with separate encoding and 9.2 with unified
encoding. These results show that learning the latent targets helps, particularly
by predicting later reasoning steps rather than reconstructing source features.

\begin{figure}[!htbp]
  \centering
  \includegraphics[width=\textwidth]{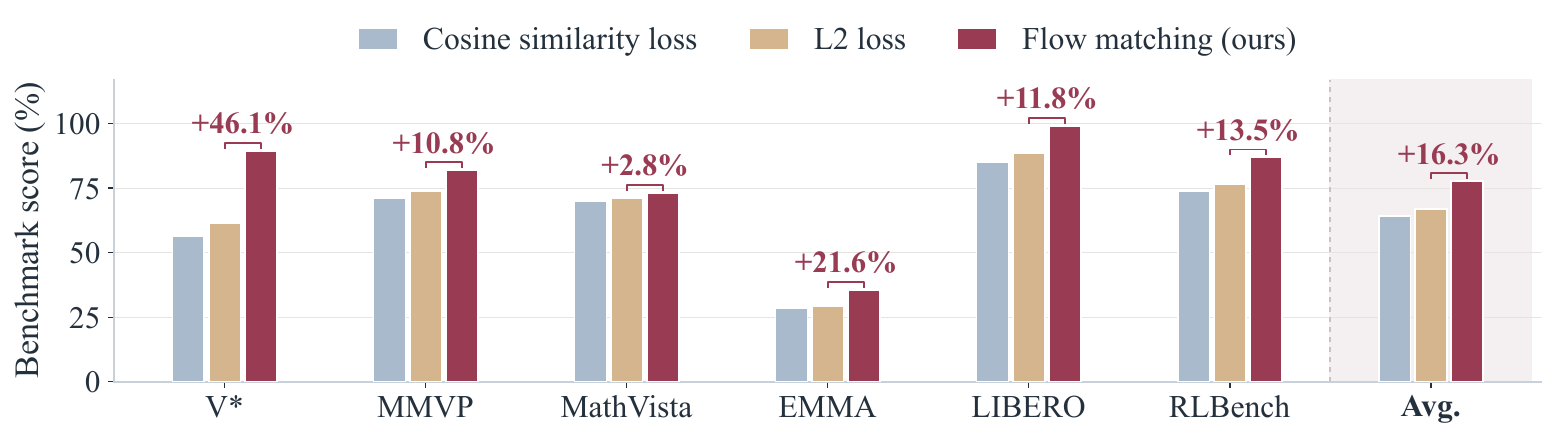}
  \caption{Comparison between different latent generation objectives.}
  \label{fig:ablation-diffusion}
\end{figure}

\paragraph{Diffusion objective.}
Finally, we test whether diffusion improves thought-token prediction\rrev{, keeping joint training of the encoder and reasoner fixed across objectives}.
We compare flow matching with direct thought-token prediction
trained using squared L2 or cosine similarity loss
(Appendix~\ref{app:ablation-protocols}). Flow matching improves the
six-benchmark average by \rrev{16.3\%} relative to L2 and \rrev{21.0\%}
relative to cosine similarity loss (Figure~\ref{fig:ablation-diffusion}).
These results show that diffusion-based thought-token prediction is an
effective alternative to direct regression for downstream reasoning and
manipulation.

%

\section{Related Work}
\label{sec:related}

\paragraph{Unified multimodal models.}
Early generalist models cast heterogeneous tasks as a shared sequence problem, spanning vision--language understanding, generation, and embodied control~\citep{wang2022ofa,lu2022unifiedio,lu2023unifiedio2,reed2022gato,mizrahi20234m,bachmann20244m21}, and large-scale vision--language models connected pretrained language models to visual inputs through learned interfaces~\citep{alayrac2022flamingo,chen2022pali,chen2023palix,huang2023kosmos1,peng2023kosmos2}.
Autoregressive any-to-any systems represent multiple modalities in a common token stream~\citep{sun2023emu,sun2023emu2,yu2023cm3leon,ge2023seed,jin2023lavit,wu2023nextgpt,zhan2024anygpt,chameleon2024,wang2024emu3}, while hybrid autoregressive--diffusion architectures unify discrete and continuous generation~\citep{tang2023codi_gen,zhou2024transfusion,xie2024showo,sun2024latentlm,wu2024janus,chen2025januspro,deng2025bagel,xu2025qwenomni,bai2025qwen3vl}.
\rev{Whereas these models unify what is read and generated, \method{} unifies the reasoning in between, representing and predicting the intermediate thoughts from every modality in a single latent space.}

\paragraph{Multimodal reasoning.}
Multimodal chain-of-thought methods supervise or elicit textual rationales from visual evidence~\citep{kang2025gflowvlm,lu2022scienceqa,zhang2023multimodalcot,rose2023visualcot,xu2024llavacot,dong2024insightv,guo2024mammothvl,yao2024mulberry,huang2025visionr1}, or externalize computation through programs, tools, image operations, spatial abstractions, and visual scratchpads~\citep{gupta2022visprog,suris2023vipergpt,yang2023mmreact,wu2023visualchatgpt,hu2023vpd,chen2024spatialvlm,hu2024visualsketchpad,zhou2024imageofthought}.
Recent work interleaves language with visual thoughts~\citep{gao2024icot,li2025zebracot,gu2025thinkmorph,li2026visualopsd}, and embodied methods guide control with language plans, visual predictions, and state or action trajectories~\citep{ahn2022saycan,huang2022innermonologue,driess2023palme,brohan2023rt2,jiang2022vima,li2023roboflamingo,kim2024openvla,li2024cogact,zhao2025cotvla,zhong2026dualcot,bai2026laravla,liu2026last0}.
Unlike methods that interleave modality-specific reasoning representations, \method{} reasons in a shared latent space and generates thought tokens through diffusion.

\paragraph{Latent reasoning.}
Latent methods internalize rationales or allocate extra computation without decoding every step~\citep{zelikman2022star,goyal2023pause,deng2023implicit,deng2024internalize,pfau2024dot,zelikman2024quietstar,yu2024system2,wang2024grokked}, compute through recurrence or continuous semantic states~\citep{dehghani2018universal,giannou2023looped,yang2023learningalgorithms,kohli2026loopthink,jeddi2026loopformer,yang2026stars,tu2026latent}, or replace language rationales with continuous thoughts~\citep{hao2024coconut,shen2025codi,xu2025softcot,xu2025softcotpp,deng2025latentsft}. Multimodal variants build latent thoughts from visual features, perceptual teachers, helper images, or task loss~\citep{kang2026scaffolding, pham2025mcout,li2025lvr,yang2025mirage,qin2025covt,wang2026monet,jeon2026valr,viveiros2026lantern,li2026livr,fan2026slvr,ray2025mulltokens,hu2026colt,chen2025ivtlr,shao2026modalmixed,chen2026lastr1}, and diffusion-based approaches model multimodal or multi-step latent trajectories~\citep{kang2025ladir,tu2026latent,he2023multimodallatent,huang2025thinkact,bai2026laravla,liu2026last0,viveiros2026holding,wu2026continuousvla,fan2026lotus,kang2026ladi}.
\newrev{Mull-Tokens~\citep{ray2025mulltokens} is the closest to our approach: although its latent tokens are shared across modalities, each token is still supervised toward the target of its corresponding step---either text tokens or frozen image-encoder features. LaST$_0$~\citep{liu2026last0} is the closest VLA counterpart but uses a separate latent format for each modality. 
\method{} instead learns a shared latent space through continuation prediction, rather than matching each step to text tokens or frozen modality-specific features.}

\section{Conclusion}
\label{sec:conclusion}

We introduced Uni-LaDiR, a unified framework for multimodal reasoning that represents all intermediate reasoning steps in a shared latent space. Its shared latent interface allows the reasoning model to predict successive steps in a common representation space. Uni-LaDiR outperforms the strongest prior methods by 7.3\% on four mathematical and logical VLM benchmarks and 6.1\% on RLBench. Notably, the advantage becomes larger as more modalities are incorporated into the shared space. These findings highlight the value of latent-space unification for enabling more effective multimodal reasoning.

\newpage


\bibliography{iclr2027_conference}

@article{lu2023unifiedio2,
  title={{Unified-IO 2}: Scaling Autoregressive Multimodal Models with Vision, Language, Audio, and Action},
  author={Lu, Jiasen and Clark, Christopher and Lee, Sangho and Zhang, Zichen and Khosla, Savya and Marten, Ryan and Hoiem, Derek and Kembhavi, Aniruddha},
  journal={arXiv preprint arXiv:2312.17172},
  year={2023}
}

@article{chameleon2024,
  title={Chameleon: Mixed-Modal Early-Fusion Foundation Models},
  author={{Chameleon Team}},
  journal={arXiv preprint arXiv:2405.09818},
  year={2024}
}

@article{zhou2024transfusion,
  title={Transfusion: Predict the Next Token and Diffuse Images with One Multi-Modal Model},
  author={Zhou, Chunting and Yu, Lili and Babu, Arun and Tirumala, Kushal and Yasunaga, Michihiro and Shamis, Leonid and Kahn, Jacob and Ma, Xuezhe and Zettlemoyer, Luke and Levy, Omer},
  journal={arXiv preprint arXiv:2408.11039},
  year={2024}
}

@article{sun2024latentlm,
  title={Multimodal Latent Language Modeling with Next-Token Diffusion},
  author={Sun, Yutao and Bao, Hangbo and Wang, Wenhui and Peng, Zhiliang and Dong, Li and Huang, Shaohan and Wang, Jianyong and Wei, Furu},
  journal={arXiv preprint arXiv:2412.08635},
  year={2024}
}

@inproceedings{wei2022cot,
  title={Chain-of-Thought Prompting Elicits Reasoning in Large Language Models},
  author={Wei, Jason and Wang, Xuezhi and Schuurmans, Dale and Bosma, Maarten and Ichter, Brian and Xia, Fei and Chi, Ed H. and Le, Quoc V. and Zhou, Denny},
  booktitle={Advances in Neural Information Processing Systems},
  year={2022}
}

@article{zhang2023multimodalcot,
  title={Multimodal Chain-of-Thought Reasoning in Language Models},
  author={Zhang, Zhuosheng and Zhang, Aston and Li, Mu and Zhao, Hai and Karypis, George and Smola, Alex},
  journal={Transactions on Machine Learning Research},
  year={2024}
}

@article{hao2024coconut,
  title={Training Large Language Models to Reason in a Continuous Latent Space},
  author={Hao, Shibo and Sukhbaatar, Sainbayar and Su, DiJia and Li, Xian and Hu, Zhiting and Weston, Jason and Tian, Yuandong},
  journal={arXiv preprint arXiv:2412.06769},
  year={2024}
}

@inproceedings{shen2025codi,
  title={{CODI}: Compressing Chain-of-Thought into Continuous Space via Self-Distillation},
  author={Shen, Zhenyi and Yan, Hanqi and Zhang, Linhai and Hu, Zhanghao and Du, Yali and He, Yulan},
  booktitle={Proceedings of the Conference on Empirical Methods in Natural Language Processing},
  year={2025}
}

@article{deng2025latentsft,
  title={{LLM} Latent Reasoning as Chain of Superposition},
  author={Deng, Jingcheng and Pang, Liang and Wei, Zihao and Xu, Shicheng and Duan, Zenghao and Xu, Kun and Song, Yang and Shen, Huawei and Cheng, Xueqi},
  journal={arXiv preprint arXiv:2510.15522},
  year={2025}
}

@article{hu2026colt,
  title={{CoLT}: Teaching Multi-Modal Models to Think with Chain of Latent Thoughts},
  author={Hu, Lianyu and Qin, Shengqian and Liao, Zeqin and Guo, Qing and Wan, Liang and Feng, Wei and Liu, Yang},
  journal={arXiv preprint arXiv:2606.31986},
  year={2026}
}

@article{kang2025ladir,
  title={{LaDiR}: Latent Diffusion Enhances {LLM}s for Text Reasoning},
  author={Kang, Haoqiang and Zhang, Yizhe and Kuang, Nikki Lijing and Majamaki, Nicklas and Jaitly, Navdeep and Ma, Yi-An and Qin, Lianhui},
  journal={arXiv preprint arXiv:2510.04573},
  year={2025}
}

@article{pham2025mcout,
  title={Multimodal Chain of Continuous Thought for Latent-Space Reasoning in Vision-Language Models},
  author={Pham, Tan-Hanh and Ngo, Chris},
  journal={arXiv preprint arXiv:2508.12587},
  year={2025}
}

@article{li2025lvr,
  title={Latent Visual Reasoning},
  author={Li, Bangzheng and Sun, Ximeng and Liu, Jiang and Wang, Ze and Wu, Jialian and Yu, Xiaodong and Chen, Hao and Barsoum, Emad and Chen, Muhao and Liu, Zicheng},
  journal={arXiv preprint arXiv:2509.24251},
  year={2025}
}

@article{bai2026laravla,
  title={Latent Reasoning {VLA}: Latent Thinking and Prediction for Vision-Language-Action Models},
  author={Bai, Shuanghao and Lyu, Jing and Zhou, Wanqi and Li, Zhe and Wang, Dakai and Xing, Lei and Zhao, Xiaoguang and Wang, Pengwei and Wang, Zhongyuan and Chi, Cheng and Chen, Badong and Zhang, Shanghang},
  journal={arXiv preprint arXiv:2602.01166},
  year={2026}
}

@article{liu2026last0,
  title={{LaST}$_0$: Latent Spatio-Temporal Chain-of-Thought for Robotic Vision-Language-Action Model},
  author={Liu, Zhuoyang and Liu, Jiaming and Chen, Hao and Yu, Jiale and Guo, Ziyu and Hou, Chengkai and Gu, Chenyang and Mi, Xiangju and Zhang, Renrui and Wu, Kun and Che, Zhengping and Tang, Jian and Heng, Pheng-Ann and Zhang, Shanghang},
  journal={arXiv preprint arXiv:2601.05248},
  year={2026}
}

@article{li2025zebracot,
  title={Zebra-CoT: A Dataset for Interleaved Vision Language Reasoning},
  author={Li, Ang and Wang, Charles and Fu, Deqing and Yue, Kaiyu and Cai, Zikui and Zhu, Wang Bill and Liu, Ollie and Guo, Peng and Neiswanger, Willie and Huang, Furong and Goldstein, Tom and Goldblum, Micah},
  journal={arXiv preprint arXiv:2507.16746},
  year={2025}
}

@article{gu2025thinkmorph,
  title={ThinkMorph: Emergent Properties in Multimodal Interleaved Chain-of-Thought Reasoning},
  author={Gu, Jiawei and Hao, Yunzhuo and Wang, Huichen Will and Li, Linjie and Shieh, Michael Qizhe and Choi, Yejin and Krishna, Ranjay and Cheng, Yu},
  journal={arXiv preprint arXiv:2510.27492},
  year={2025}
}

@inproceedings{viveiros2026ilvr,
  title={Interleaved Latent Visual Reasoning with Selective Perceptual Modeling},
  author={Dong, Shuai and Wang, Siyuan and Liu, Xingyu and Li, Chenglin and Hou, Haowen and Wei, Zhongyu},
  booktitle={Proceedings of the 64th Annual Meeting of the Association for Computational Linguistics (Volume 1: Long Papers)},
  pages={29316--29335},
  year={2026}
}

@article{li2026visualopsd,
  title={Visual-OPSD: Cross-Modal On-Policy Self-Distillation for Efficient Unified Multimodal Reasoning},
  author={Li, Pengyu and Gao, Zhitao and Zhang, Lingling and Huang, Muye and Li, Yuanming and Yang, Zesheng and Xu, Fangzhi and Liu, Jun},
  journal={arXiv preprint arXiv:2606.18974},
  year={2026}
}

@article{bai2025qwen3vl,
  title={{Qwen3-VL} Technical Report},
  author={Bai, Shuai and Cai, Yuxuan and Chen, Ruizhe and Chen, Keqin and Chen, Xionghui and Cheng, Zesen and others},
  journal={arXiv preprint arXiv:2511.21631},
  year={2025}
}

@article{deng2025bagel,
  title={Emerging Properties in Unified Multimodal Pretraining},
  author={Deng, Chaorui and Zhu, Deyao and Li, Kunchang and Gou, Chenhui and Li, Feng and Wang, Zeyu and Zhong, Shu and Yu, Weihao and Nie, Xiaonan and Song, Ziang and Shi, Guang and Fan, Haoqi},
  journal={arXiv preprint arXiv:2505.14683},
  year={2025}
}

@article{chen2025januspro,
  title={Janus-Pro: Unified Multimodal Understanding and Generation with Data and Model Scaling},
  author={Chen, Xiaokang and Wu, Zhiyu and Liu, Xingchao and Pan, Zizheng and Liu, Wen and Xie, Zhenda and Yu, Xingkai and Ruan, Chong},
  journal={arXiv preprint arXiv:2501.17811},
  year={2025}
}

@inproceedings{masry2022chartqa,
  title={ChartQA: A Benchmark for Question Answering about Charts with Visual and Logical Reasoning},
  author={Masry, Ahmed and Long, Do Xuan and Tan, Jia Qing and Joty, Shafiq and Hoque, Enamul},
  booktitle={Findings of the Association for Computational Linguistics: ACL 2022},
  year={2022}
}

@inproceedings{wu2024vstar,
  title={{V*}: Guided Visual Search as a Core Mechanism in Multimodal {LLM}s},
  author={Wu, Penghao and Xie, Saining},
  booktitle={Proceedings of the IEEE/CVF Conference on Computer Vision and Pattern Recognition},
  year={2024}
}

@inproceedings{fu2024blink,
  title={{BLINK}: Multimodal Large Language Models Can See but Not Perceive},
  author={Fu, Xingyu and Hu, Yushi and Li, Bangzheng and Feng, Yu and Wang, Haoyu and Lin, Xudong and Roth, Dan and Smith, Noah A. and Ma, Wei-Chiu and Krishna, Ranjay},
  booktitle={European Conference on Computer Vision},
  year={2024}
}

@inproceedings{tong2024mmvp,
  title={Eyes Wide Shut? Exploring the Visual Shortcomings of Multimodal {LLM}s},
  author={Tong, Shengbang and Liu, Zhuang and Zhai, Yuexiang and Ma, Yi and LeCun, Yann and Xie, Saining},
  booktitle={Proceedings of the IEEE/CVF Conference on Computer Vision and Pattern Recognition},
  year={2024}
}

@article{ray2024sat,
  title={{SAT}: Dynamic Spatial Aptitude Training for Multimodal Language Models},
  author={Ray, Arijit and Duan, Jiafei and Brown, Ellis and Tan, Reuben and Bashkirova, Dina and Hendrix, Rose and Ehsani, Kiana and Kembhavi, Aniruddha and Plummer, Bryan A. and Krishna, Ranjay and Zeng, Kuo-Hao and Saenko, Kate},
  journal={arXiv preprint arXiv:2412.07755},
  year={2024}
}

@inproceedings{tong2024cambrian,
  title={Cambrian-1: A Fully Open, Vision-Centric Exploration of Multimodal {LLM}s},
  author={Tong, Shengbang and Brown, Ellis and Wu, Penghao and Woo, Sanghyun and Middepogu, Manoj and Akula, Sai Charitha and Yang, Jihan and Yang, Shusheng and Iyer, Adithya and Pan, Xichen and Wang, Austin and Fergus, Rob and LeCun, Yann and Xie, Saining},
  booktitle={Advances in Neural Information Processing Systems},
  year={2024}
}

@article{song2025visualpuzzles,
  title={VisualPuzzles: Decoupling Multimodal Reasoning Evaluation from Domain Knowledge},
  author={Song, Yueqi and Ou, Tianyue and Kong, Yibo and Li, Zecheng and Neubig, Graham and Yue, Xiang},
  journal={arXiv preprint arXiv:2504.10342},
  year={2025}
}

@inproceedings{liu2024libero,
  title={{LIBERO}: Benchmarking Knowledge Transfer for Lifelong Robot Learning},
  author={Liu, Bo and Zhu, Yifeng and Gao, Chongkai and Feng, Yihao and Liu, Qiang and Zhu, Yuke and Stone, Peter},
  booktitle={Advances in Neural Information Processing Systems},
  volume={36},
  year={2023}
}

@article{james2020rlbench,
  title={{RLBench}: The Robot Learning Benchmark \& Learning Environment},
  author={James, Stephen and Ma, Zicong and Arrojo, David Rovick and Davison, Andrew J.},
  journal={IEEE Robotics and Automation Letters},
  volume={5},
  number={2},
  pages={3019--3026},
  year={2020}
}

@inproceedings{kim2024openvla,
  title={OpenVLA: An Open-Source Vision-Language-Action Model},
  author={Kim, Moo Jin and Pertsch, Karl and Karamcheti, Siddharth and Xiao, Ted and Balakrishna, Ashwin and Nair, Suraj and Rafailov, Rafael and Foster, Ethan and Lam, Grace and Sanketi, Pannag and others},
  booktitle={Conference on Robot Learning},
  year={2024}
}

@article{qu2025spatialvla,
  title={SpatialVLA: Exploring Spatial Representations for Visual-Language-Action Model},
  author={Qu, Delin and Song, Haoming and Chen, Qizhi and Yao, Yuanqi and Ye, Xinyi and Ding, Yan and Wang, Zhigang and Gu, Jiayuan and Zhao, Bin and Wang, Dong and Li, Xuelong},
  journal={arXiv preprint arXiv:2501.15830},
  year={2025}
}

@article{li2024cogact,
  title={CogACT: A Foundational Vision-Language-Action Model for Synergizing Cognition and Action in Robotic Manipulation},
  author={Li, Qixiu and Liang, Yaobo and Wang, Zeyu and Luo, Lin and Chen, Xi and Liao, Mozheng and Wei, Fangyun and Deng, Yu and Xu, Sicheng and Zhang, Yizhong and others},
  journal={arXiv preprint arXiv:2411.19650},
  year={2024}
}

@article{zhao2025cotvla,
  title={CoT-VLA: Visual Chain-of-Thought Reasoning for Vision-Language-Action Models},
  author={Zhao, Qingqing and Lu, Yao and Kim, Moo Jin and Fu, Zipeng and Zhang, Zhuoyang and Wu, Yecheng and Li, Zhaoshuo and Ma, Qianli and Han, Song and Finn, Chelsea and Handa, Ankur and Liu, Ming-Yu and Xiang, Donglai and Wetzstein, Gordon and Lin, Tsung-Yi},
  journal={arXiv preprint arXiv:2503.22020},
  year={2025}
}

@article{pi05,
  title={$\pi_{0.5}$: A Vision-Language-Action Model with Open-World Generalization},
  author={{Physical Intelligence} and Black, Kevin and Brown, Noah and Darpinian, James and Dhabalia, Karan and Driess, Danny and Esmail, Adnan and Equi, Michael and Finn, Chelsea and others},
  journal={arXiv preprint arXiv:2504.16054},
  year={2025}
}

@article{kim2025openvlaoft,
  title={Fine-Tuning Vision-Language-Action Models: Optimizing Speed and Success},
  author={Kim, Moo Jin and Finn, Chelsea and Liang, Percy},
  journal={arXiv preprint arXiv:2502.19645},
  year={2025}
}

@article{liu2025hybridvla,
  title={HybridVLA: Collaborative Diffusion and Autoregression in a Unified Vision-Language-Action Model},
  author={Liu, Jiaming and Chen, Hao and An, Pengju and Liu, Zhuoyang and Zhang, Renrui and Gu, Chenyang and Li, Xiaoqi and Guo, Ziyu and Chen, Sixiang and Liu, Mengzhen and others},
  journal={arXiv preprint arXiv:2503.10631},
  year={2025}
}

@inproceedings{chen2025fisvla,
  title={Fast-in-Slow: A Dual-System VLA Model Unifying Fast Manipulation within Slow Reasoning},
  author={Chen, Hao and Liu, Jiaming and Gu, Chenyang and Liu, Zhuoyang and Zhang, Renrui and Li, Xiaoqi and He, Xiao and Guo, Yandong and Fu, Chi-Wing and Zhang, Shanghang and Heng, Pheng-Ann},
  booktitle={Advances in Neural Information Processing Systems},
  year={2025}
}

@article{zheng2025deepeyes,
  title={DeepEyes: Incentivizing Thinking with Images via Reinforcement Learning},
  author={Zheng, Ziwei and Yang, Michael and Hong, Jack and Zhao, Chenxiao and Xu, Guohai and Yang, Le and Shen, Chao and Yu, Xing},
  journal={arXiv preprint arXiv:2505.14362},
  year={2025}
}

@article{jeon2026valr,
  title={Vision-aligned Latent Reasoning for Multi-modal Large Language Model},
  author={Jeon, Byungwoo and Jeong, Yoonwoo and Lee, Hyunseok and Cho, Minsu and Shin, Jinwoo},
  journal={arXiv preprint arXiv:2602.04476},
  year={2026}
}

@article{viveiros2026lantern,
  title={{LanteRn}: Latent Visual Structured Reasoning},
  author={Viveiros, Andr\'e G. and Gon\c{c}alves, Nuno and Lindemann, Matthias and Martins, Andr\'e},
  journal={arXiv preprint arXiv:2603.25629},
  year={2026}
}

@article{fan2026slvr,
  title={Semantic-Enriched Latent Visual Reasoning},
  author={Xu, Tianrun and Sun, Yue and Wang, Qixun and Lu, Jingyi and Wang, Yuan and Zhang, Tianren and Guo, Longteng and Rao, Fengyun and Lyu, Jing and Chen, Feng and Liu, Jing},
  journal={arXiv preprint arXiv:2605.19342},
  year={2026}
}

@article{ray2025mulltokens,
  title={Mull-Tokens: Modality-Agnostic Latent Thinking},
  author={Ray, Arijit and Abdelkader, Ahmed and Mao, Chengzhi and Plummer, Bryan A. and Saenko, Kate and Krishna, Ranjay and Guibas, Leonidas and Chu, Wen-Sheng},
  journal={arXiv preprint arXiv:2512.10941},
  year={2025}
}

@article{yang2025mirage,
  title={Machine Mental Imagery: Empower Multimodal Reasoning with Latent Visual Tokens},
  author={Yang, Zeyuan and Yu, Xueyang and Chen, Delin and Shen, Maohao and Gan, Chuang},
  journal={arXiv preprint arXiv:2506.17218},
  year={2025}
}

@article{qin2025covt,
  title={Chain-of-Visual-Thought: Teaching {VLM}s to See and Think Better with Continuous Visual Tokens},
  author={Qin, Yiming and Wei, Bomin and Ge, Jiaxin and Kallidromitis, Konstantinos and Fu, Stephanie and Darrell, Trevor and Wang, Xudong},
  journal={arXiv preprint arXiv:2511.19418},
  year={2025}
}

@inproceedings{wang2026monet,
  title={Monet: Reasoning in Latent Visual Space Beyond Image and Language},
  author={Wang, Qixun and Shi, Yang and Wang, Yifei and Zhang, Yuanxing and Wan, Pengfei and Gai, Kun and Ying, Xianghua and Wang, Yisen},
  booktitle={Proceedings of the IEEE/CVF Conference on Computer Vision and Pattern Recognition},
  pages={12030--12040},
  year={2026}
}

@inproceedings{li2026livr,
  title={Latent Implicit Visual Reasoning},
  author={Li, Kelvin and Shang, Chuyi and Karlinsky, Leonid and Feris, Rogerio and Darrell, Trevor and Herzig, Roei},
  booktitle={Proceedings of the IEEE/CVF Conference on Computer Vision and Pattern Recognition},
  pages={33457--33466},
  year={2026}
}

@inproceedings{lu2024mathvista,
  title={MathVista: Evaluating Mathematical Reasoning of Foundation Models in Visual Contexts},
  author={Lu, Pan and Bansal, Hritik and Xia, Tony and Liu, Jiacheng and Li, Chunyuan and Hajishirzi, Hannaneh and Cheng, Hao and Chang, Kai-Wei and Galley, Michel and Gao, Jianfeng},
  booktitle={International Conference on Learning Representations},
  year={2024}
}

@inproceedings{wang2024mathvision,
  title={Measuring Multimodal Mathematical Reasoning with {MATH-Vision} Dataset},
  author={Wang, Ke and Pan, Junting and Shi, Weikang and Lu, Zimu and Ren, Houxing and Zhou, Aojun and Zhan, Mingjie and Li, Hongsheng},
  booktitle={Advances in Neural Information Processing Systems Datasets and Benchmarks Track},
  year={2024}
}

@article{xu2025visulogic,
  title={VisuLogic: A Benchmark for Evaluating Visual Reasoning in Multi-Modal Large Language Models},
  author={Xu, Weiye and Wang, Jiahao and Wang, Weiyun and Chen, Zhe and Zhou, Wengang and Yang, Aijun and Lu, Lewei and Li, Houqiang and Wang, Xiaohua and Zhu, Xizhou and Wang, Wenhai and Dai, Jifeng and Zhu, Jinguo},
  journal={arXiv preprint arXiv:2504.15279},
  year={2025}
}

@inproceedings{hao2025emma,
  title={Can {MLLM}s Reason in Multimodality? {EMMA}: An Enhanced Multimodal Reasoning Benchmark},
  author={Hao, Yunzhuo and Gu, Jiawei and Wang, Huichen Will and Li, Linjie and Yang, Zhengyuan and Wang, Lijuan and Cheng, Yu},
  booktitle={International Conference on Machine Learning},
  year={2025}
}

@inproceedings{wang2022ofa,title={{OFA}: Unifying Architectures, Tasks, and Modalities Through a Simple Sequence-to-Sequence Learning Framework},author={Wang, Peng and others},booktitle={International Conference on Machine Learning},year={2022}}

@article{lu2022unifiedio,title={Unified-IO: A Unified Model for Vision, Language, and Multi-Modal Tasks},author={Lu, Jiasen and Clark, Christopher and Zellers, Rowan and Mottaghi, Roozbeh and Kembhavi, Aniruddha},journal={arXiv preprint arXiv:2206.08916},year={2022}}

@article{reed2022gato,title={A Generalist Agent},author={Reed, Scott and Zolna, Konrad and Parisotto, Emilio and others},journal={Transactions on Machine Learning Research},year={2022}}

@article{mizrahi20234m,title={{4M}: Massively Multimodal Masked Modeling},author={Mizrahi, David and Bachmann, Roman and Kar, Oguzhan Fatih and Yeo, Teresa and Gao, Mingfei and Dehghan, Afshin and Zamir, Amir},journal={arXiv preprint arXiv:2312.06647},year={2023}}

@article{bachmann20244m21,title={{4M-21}: An Any-to-Any Vision Model for Tens of Tasks and Modalities},author={Bachmann, Roman and Kar, O{\u{g}}uzhan Fatih and Mizrahi, David and others},journal={arXiv preprint arXiv:2406.09406},year={2024}}

@article{alayrac2022flamingo,title={Flamingo: A Visual Language Model for Few-Shot Learning},author={Alayrac, Jean-Baptiste and Donahue, Jeff and Luc, Pauline and others},journal={Advances in Neural Information Processing Systems},year={2022}}

@article{chen2022pali,title={{PaLI}: A Jointly-Scaled Multilingual Language-Image Model},author={Chen, Xi and Wang, Xiao and Changpinyo, Soravit and others},journal={arXiv preprint arXiv:2209.06794},year={2022}}

@article{chen2023palix,title={{PaLI-X}: On Scaling Up a Multilingual Vision and Language Model},author={Chen, Xi and others},journal={arXiv preprint arXiv:2305.18565},year={2023}}

@article{huang2023kosmos1,title={Language Is Not All You Need: Aligning Perception with Language Models},author={Huang, Shaohan and Dong, Li and Wang, Wenhui and others},journal={arXiv preprint arXiv:2302.14045},year={2023}}

@inproceedings{peng2023kosmos2,title={{Kosmos-2}: Grounding Multimodal Large Language Models to the World},author={Peng, Zhiliang and Wang, Wenhui and Dong, Li and others},booktitle={International Conference on Learning Representations},year={2024}}

@inproceedings{sun2023emu,title={{Emu}: Generative Pretraining in Multimodality},author={Sun, Quan and Yu, Qiying and Cui, Yufeng and Zhang, Fan and Zhang, Xiaosong and Wang, Yueze and Gao, Hongcheng and Liu, Jingjing and Huang, Tiejun and Wang, Xinlong},booktitle={International Conference on Learning Representations},year={2024}}

@inproceedings{sun2023emu2,title={Generative Multimodal Models are In-Context Learners},author={Sun, Quan and Cui, Yufeng and Zhang, Xiaosong and others},booktitle={Proceedings of the IEEE/CVF Conference on Computer Vision and Pattern Recognition},year={2024}}

@article{yu2023cm3leon,title={Scaling Autoregressive Multi-Modal Models: Pretraining and Instruction Tuning},author={Yu, Lili and Shi, Bowen and Pasunuru, Ramakanth and others},journal={arXiv preprint arXiv:2309.02591},year={2023}}

@article{ge2023seed,title={Planting a {SEED} of Vision in Large Language Model},author={Ge, Yuying and Ge, Yixiao and Zeng, Ziyun and others},journal={arXiv preprint arXiv:2307.08041},year={2023}}

@inproceedings{jin2023lavit,title={Unified Language-Vision Pretraining in {LLM} with Dynamic Discrete Visual Tokenization},author={Jin, Yang and Xu, Kun and Xu, Kun and others},booktitle={International Conference on Learning Representations},year={2024}}

@article{wu2023nextgpt,title={{NExT-GPT}: Any-to-Any Multimodal LLM},author={Wu, Shengqiong and Fei, Hao and Qu, Leigang and Ji, Wei and Chua, Tat-Seng},journal={arXiv preprint arXiv:2309.05519},year={2023}}

@article{zhan2024anygpt,title={{AnyGPT}: Unified Multimodal LLM with Discrete Sequence Modeling},author={Zhan, Jun and Dai, Junqi and Ye, Jiasheng and others},journal={arXiv preprint arXiv:2402.12226},year={2024}}

@article{wang2024emu3,title={{Emu3}: Next-Token Prediction is All You Need},author={Wang, Xinlong and Zhang, Xiaosong and Luo, Zhengxiong and others},journal={arXiv preprint arXiv:2409.18869},year={2024}}

@article{tang2023codi_gen,title={Any-to-Any Generation via Composable Diffusion},author={Tang, Zineng and Yang, Ziyi and Zhu, Chenguang and Zeng, Michael and Bansal, Mohit},journal={arXiv preprint arXiv:2305.11846},year={2023}}

@article{xie2024showo,title={{Show-o}: One Single Transformer to Unify Multimodal Understanding and Generation},author={Xie, Jinheng and Mao, Weijia and Bai, Zechen and others},journal={arXiv preprint arXiv:2408.12528},year={2024}}

@inproceedings{wu2024janus,title={Janus: Decoupling Visual Encoding for Unified Multimodal Understanding and Generation},author={Wu, Chengyue and Chen, Xiaokang and Wu, Zhiyu and others},booktitle={Proceedings of the IEEE/CVF Conference on Computer Vision and Pattern Recognition},year={2025}}

@article{xu2025qwenomni,title={{Qwen2.5-Omni} Technical Report},author={Xu, Jin and others},journal={arXiv preprint arXiv:2503.20215},year={2025}}

@article{lu2022scienceqa,title={Learn to Explain: Multimodal Reasoning via Thought Chains for Science Question Answering},author={Lu, Pan and Mishra, Swaroop and Xia, Tanglin and Qiu, Liang and Chang, Kai-Wei and Zhu, Song-Chun and Tafjord, Oyvind and Clark, Peter and Kalyan, Ashwin},journal={Advances in Neural Information Processing Systems},year={2022}}

@article{rose2023visualcot,title={Visual Chain of Thought: Bridging Logical Gaps with Multimodal Infillings},author={Rose, Daniel and others},journal={arXiv preprint arXiv:2305.02317},year={2023}}

@article{xu2024llavacot,title={{LLaVA-CoT}: Let Vision Language Models Reason Step-by-Step},author={Xu, Guowei and others},journal={arXiv preprint arXiv:2411.10440},year={2024}}

@article{dong2024insightv,title={{Insight-V}: Exploring Long-Chain Visual Reasoning with Multimodal Large Language Models},author={Dong, Yuhao and others},journal={arXiv preprint arXiv:2411.14432},year={2024}}

@article{guo2024mammothvl,title={{MAmmoTH-VL}: Eliciting Multimodal Reasoning with Instruction Tuning at Scale},author={Guo, Jarvis and others},journal={arXiv preprint arXiv:2412.05237},year={2024}}

@article{yao2024mulberry,title={Mulberry: Empowering MLLM with O1-like Reasoning and Reflection via Collective Monte Carlo Tree Search},author={Yao, Huanjin and others},journal={arXiv preprint arXiv:2412.18319},year={2024}}

@article{huang2025visionr1,title={{Vision-R1}: Incentivizing Reasoning Capability in Multimodal Large Language Models},author={Huang, Wenxuan and others},journal={arXiv preprint arXiv:2503.06749},year={2025}}

@article{su2025pixelreasoner,
  title={Pixel Reasoner: Incentivizing Pixel-Space Reasoning with Curiosity-Driven Reinforcement Learning},
  author={Wang, Haozhe and Su, Alex and Ren, Weiming and Lin, Fangzhen and Chen, Wenhu},
  journal={arXiv preprint arXiv:2505.15966},
  year={2025}
}

@article{gupta2022visprog,title={Visual Programming: Compositional Visual Reasoning Without Training},author={Gupta, Tanmay and Kembhavi, Aniruddha},journal={arXiv preprint arXiv:2211.11559},year={2022}}

@article{suris2023vipergpt,title={{ViperGPT}: Visual Inference via Python Execution for Reasoning},author={Suris, Didac and Menon, Sachit and Vondrick, Carl},journal={arXiv preprint arXiv:2303.08128},year={2023}}

@article{yang2023mmreact,title={{MM-REACT}: Prompting ChatGPT for Multimodal Reasoning and Action},author={Yang, Zhengyuan and Li, Linjie and Wang, Jianfeng and Lin, Kevin and Azarnasab, Ehsan and Ahmed, Faisal and Liu, Zicheng and Liu, Ce and Zeng, Michael and Wang, Lijuan},journal={arXiv preprint arXiv:2303.11381},year={2023}}

@article{wu2023visualchatgpt,title={Visual ChatGPT: Talking, Drawing and Editing with Visual Foundation Models},author={Wu, Chenfei and Yin, Shengming and Qi, Weizhen and others},journal={arXiv preprint arXiv:2303.04671},year={2023}}

@inproceedings{hu2023vpd,title={Visual Program Distillation: Distilling Tools and Programmatic Reasoning into Vision-Language Models},author={Hu, Yushi and others},booktitle={Proceedings of the IEEE/CVF Conference on Computer Vision and Pattern Recognition},year={2024}}

@article{chen2024spatialvlm,title={SpatialVLM: Endowing Vision-Language Models with Spatial Reasoning Capabilities},author={Chen, Boyuan and others},journal={arXiv preprint arXiv:2401.12168},year={2024}}

@inproceedings{hu2024visualsketchpad,title={Visual Sketchpad: Sketching as a Visual Chain of Thought for Multimodal Language Models},author={Hu, Yushi and Shi, Weijia and Fu, Xingyu and others},booktitle={Advances in Neural Information Processing Systems},year={2024}}

@article{zhou2024imageofthought,title={Image-of-Thought Prompting for Visual Reasoning Refinement in Multimodal Large Language Models},author={Zhou, Qiji and others},journal={arXiv preprint arXiv:2405.13872},year={2024}}

@article{gao2024icot,title={Interleaved-Modal Chain-of-Thought},author={Gao, Jun and Li, Yongqi and Cao, Ziqiang and Li, Wenjie},journal={arXiv preprint arXiv:2411.19488},year={2024}}

@article{li2025mvot,title={Imagine while Reasoning in Space: Multimodal Visualization-of-Thought},author={Li, Chengzu and Wu, Wenshan and Zhang, Huanyu and Xia, Yan and Mao, Shaoguang and Dong, Li and Vuli\'c, Ivan and Wei, Furu},journal={arXiv preprint arXiv:2501.07542},year={2025}}

@article{qin2025unicot,title={Uni-{CoT}: Towards Unified Chain-of-Thought Reasoning Across Text and Vision},author={Qin, Luozheng and Gong, Jia and Sun, Yuqing and Li, Tianjiao and Yang, Mengping and Yang, Xiaomeng and Qu, Chao and Tan, Zhiyu and Li, Hao},journal={arXiv preprint arXiv:2508.05606},year={2025}}

@article{shao2026modalmixed,
  title={Learning Modal-Mixed Chain-of-Thought Reasoning with Latent Embeddings},
  author={Shao, Yifei and Zhou, Kun and Xu, Ziming and Quamar, Mohammad Atif and Hao, Shibo and Wang, Zhen and Hu, Zhiting and Huang, Biwei},
  journal={arXiv preprint arXiv:2602.00574},
  year={2026}
}

@inproceedings{ahn2022saycan,title={Do As I Can, Not As I Say: Grounding Language in Robotic Affordances},author={Ahn, Michael and Brohan, Anthony and Brown, Noah and others},booktitle={Conference on Robot Learning},year={2022}}

@inproceedings{huang2022innermonologue,title={Inner Monologue: Embodied Reasoning through Planning with Language Models},author={Huang, Wenlong and Xia, Fei and Xiao, Ted and others},booktitle={Conference on Robot Learning},year={2022}}

@inproceedings{driess2023palme,title={{PaLM-E}: An Embodied Multimodal Language Model},author={Driess, Danny and Xia, Fei and Sajjadi, Mehdi S. M. and others},booktitle={International Conference on Machine Learning},year={2023}}

@inproceedings{brohan2023rt2,title={{RT-2}: Vision-Language-Action Models Transfer Web Knowledge to Robotic Control},author={Brohan, Anthony and Brown, Noah and Carbajal, Justice and others},booktitle={Conference on Robot Learning},year={2023}}

@article{jiang2022vima,title={{VIMA}: General Robot Manipulation with Multimodal Prompts},author={Jiang, Yunfan and Gupta, Agrim and Zhang, Zichen and others},journal={arXiv preprint arXiv:2210.03094},year={2022}}

@inproceedings{li2023roboflamingo,title={Vision-Language Foundation Models as Effective Robot Imitators},author={Li, Xinghang and Liu, Minghuan and Zhang, Hanbo and others},booktitle={International Conference on Learning Representations},year={2024}}

@article{zhong2026dualcot,
  title={{DualCoT-VLA}: Visual-Linguistic Chain of Thought via Parallel Reasoning for Vision-Language-Action Models},
  author={Zhong, Zhide and Li, Junfeng and He, Junjie and Yan, Haodong and Gong, Xin and Zhao, Guanyi and Cai, Yingjie and Gao, Jiantao and Yan, Xu and Liu, Bingbing and Chen, Yingcong and Yang, Liuqing and Li, Haoang},
  journal={arXiv preprint arXiv:2603.22280},
  year={2026}
}

@article{zelikman2022star,title={{STaR}: Bootstrapping Reasoning With Reasoning},author={Zelikman, Eric and Wu, Yuhuai and Mu, Jesse and Goodman, Noah D.},journal={Advances in Neural Information Processing Systems},year={2022}}

@inproceedings{goyal2023pause,title={Think Before You Speak: Training Language Models With Pause Tokens},author={Goyal, Sachin and Ji, Ziwei and Rawat, Ankit Singh and others},booktitle={International Conference on Learning Representations},year={2024}}

@article{deng2023implicit,title={Implicit Chain of Thought Reasoning via Knowledge Distillation},author={Deng, Yuntian and others},journal={arXiv preprint arXiv:2311.01460},year={2023}}

@article{deng2024internalize,title={From Explicit CoT to Implicit CoT: Learning to Internalize CoT Step by Step},author={Deng, Yuntian and others},journal={arXiv preprint arXiv:2405.14838},year={2024}}

@article{pfau2024dot,title={Let's Think Dot by Dot: Hidden Computation in Transformer Language Models},author={Pfau, Jacob and Merrill, William and Bowman, Samuel R.},journal={arXiv preprint arXiv:2404.15758},year={2024}}

@article{zelikman2024quietstar,title={Quiet-STaR: Language Models Can Teach Themselves to Think Before Speaking},author={Zelikman, Eric and Harik, Georges and Shao, Yijia and others},journal={arXiv preprint arXiv:2403.09629},year={2024}}

@article{yu2024system2,title={Distilling System 2 into System 1},author={Yu, Ping and others},journal={arXiv preprint arXiv:2407.06023},year={2024}}

@article{wang2024grokked,title={Grokked Transformers are Implicit Reasoners: A Mechanistic Journey to the Edge of Generalization},author={Wang, Boshi and others},journal={arXiv preprint arXiv:2405.15071},year={2024}}

@inproceedings{dehghani2018universal,title={Universal Transformers},author={Dehghani, Mostafa and Gouws, Stephan and Vinyals, Oriol and others},booktitle={International Conference on Learning Representations},year={2019}}

@article{giannou2023looped,title={Looped Transformers as Programmable Computers},author={Giannou, Angeliki and Rajput, Shashank and Sohn, Jy-Yong and Lee, Kangwook and Lee, Jason D. and Papailiopoulos, Dimitris},journal={arXiv preprint arXiv:2301.13196},year={2023}}

@inproceedings{yang2023learningalgorithms,title={Looped Transformers are Better at Learning Learning Algorithms},author={Yang, Liu and others},booktitle={International Conference on Learning Representations},year={2024}}

@article{xu2025softcot,title={SoftCoT: Soft Chain-of-Thought for Efficient Reasoning with LLMs},author={Xu, Yige and others},journal={arXiv preprint arXiv:2502.12134},year={2025}}

@article{xu2025softcotpp,title={SoftCoT++: Test-Time Scaling with Soft Chain-of-Thought Reasoning},author={Xu, Yige and others},journal={arXiv preprint arXiv:2505.11484},year={2025}}

@article{he2023multimodallatent,title={Multi-modal Latent Space Learning for Chain-of-Thought Reasoning in Language Models},author={He, Liqi and others},journal={arXiv preprint arXiv:2312.08762},year={2023}}

@inproceedings{huang2025thinkact,title={ThinkAct: Vision-Language-Action Reasoning via Reinforced Visual Latent Planning},author={Huang, Chi-Pin and others},booktitle={Advances in Neural Information Processing Systems},year={2025}}

@article{kohli2026loopthink,title={Loop, Think, \& Generalize: Implicit Reasoning in Recurrent-Depth Transformers},author={Kohli, Harsh and others},journal={arXiv preprint arXiv:2604.07822},year={2026}}

@article{jeddi2026loopformer,title={{LoopFormer}: Elastic-Depth Looped Transformers for Latent Reasoning via Shortcut Modulation},author={Jeddi, Ahmadreza and others},journal={arXiv preprint arXiv:2602.11451},year={2026}}

@article{yang2026stars,title={Stabilizing Recurrent Dynamics for Test-Time Scalable Latent Reasoning in Looped Language Models},author={Yang, Xiao-Wen and others},journal={arXiv preprint arXiv:2605.26733},year={2026}}

@article{viveiros2026holding,title={What's Holding Back Latent Visual Reasoning?},author={Viveiros, Andr\'e G. and others},journal={arXiv preprint arXiv:2605.18445},year={2026}}

@article{cui2026latentsupervision,
  title={How Do Latent Reasoning Methods Perform Under Weak and Strong Supervision?},
  author={Cui, Yingqian and Dai, Zhenwei and He, Bing and Shi, Zhan and Liu, Hui and Sun, Rui and Liu, Zhiji and Xing, Yue and Tang, Jiliang and Dumoulin, Benoit},
  journal={arXiv preprint arXiv:2602.22441},
  year={2026}
}

@article{wu2026continuousvla,title={Continuous Reasoning for Vision-Language-Action},author={Wu, Yueh-Hua and others},journal={arXiv preprint arXiv:2606.00229},year={2026}}

@article{fan2026lotus,title={Bridging the Gap Between Latent and Explicit Reasoning with Looped Transformers},author={Fan, Ying and others},journal={arXiv preprint arXiv:2606.31779},year={2026}}

@article{cen2025worldvla,
  title={WorldVLA: Towards Autoregressive Action World Model},
  author={Cen, Jun and others},
  journal={arXiv preprint arXiv:2506.21539},
  year={2025}
}

@article{wang2025univla,
  title={UniVLA: Learning to Act Anywhere with Task-centric Latent Actions},
  author={Bu, Qingwen and Yang, Yanting and Cai, Jisong and Gao, Shenyuan and Ren, Guanghui and Yao, Maoqing and Luo, Ping and Li, Hongyang},
  journal={arXiv preprint arXiv:2505.06111},
  year={2025}
}

@article{zhong2025flowvla,
  title={{FlowVLA}: Visual Chain of Thought-based Motion Reasoning for Vision-Language-Action Models},
  author={Zhong, Zhide and Yan, Haodong and Li, Junfeng and Liu, Xiangchen and Gong, Xin and Zhang, Tianran and Song, Wenxuan and Chen, Jiayi and Zheng, Xinhu and Wang, Hesheng and Li, Haoang},
  journal={arXiv preprint arXiv:2508.18269},
  year={2025}
}

@article{xu2026futurevla,
  title={{FutureVLA}: Joint Visuomotor Prediction for Vision-Language-Action Model},
  author={Xu, Xiaoxu and Li, Hao and Ye, Jinhui and Chen, Yilun and Zeng, Jia and Chen, Xinyi and Xu, Linning and Lin, Dahua and Li, Weixin and Pang, Jiangmiao},
  journal={arXiv preprint arXiv:2603.10712},
  year={2026}
}

@article{li2026consisvla,
  title={ConsisVLA-4D: Advancing Spatiotemporal Consistency in Efficient 3D-Perception and 4D-Reasoning for Robotic Manipulation},
  author={Li, Wei and Liu, Jizhihui and Yixing, Li and Tong, Junwen and Shao, Rui and Nie, Liqiang},
  journal={arXiv preprint arXiv:2605.05126},
  year={2026}
}

@article{lei2026avavla,
  title={Think Less, Act Early: Reinforced Latent Reasoning with Early Exit in Vision-Language-Action Models},
  author={Lei, Dianqiao and Shan, Lianlei},
  journal={arXiv preprint arXiv:2606.15099},
  year={2026}
}

@article{yang2026pearlvla,
  title={PearlVLA: Progressive Embodied Action-Plan Refinement in Latent Space},
  author={Yang, Bochen and Shan, Lianlei},
  journal={arXiv preprint arXiv:2606.17924},
  year={2026}
}

@inproceedings{ho2020ddpm,
  title={Denoising Diffusion Probabilistic Models},
  author={Ho, Jonathan and Jain, Ajay and Abbeel, Pieter},
  booktitle={Advances in Neural Information Processing Systems},
  volume={33},
  year={2020},
  url={https://arxiv.org/abs/2006.11239}
}

@inproceedings{lipman2023flow,
  title={Flow Matching for Generative Modeling},
  author={Lipman, Yaron and Chen, Ricky T. Q. and Ben-Hamu, Heli and Nickel, Maximilian and Le, Matt},
  booktitle={International Conference on Learning Representations},
  year={2023},
  url={https://openreview.net/forum?id=PqvMRDCJT9t}
}

@inproceedings{chi2023diffusionpolicy,
  title={Diffusion Policy: Visuomotor Policy Learning via Action Diffusion},
  author={Chi, Cheng and Feng, Siyuan and Du, Yilun and Xu, Zhenjia and Cousineau, Eric and Burchfiel, Benjamin C. M. and Song, Shuran},
  booktitle={Proceedings of Robotics: Science and Systems},
  year={2023},
  doi={10.15607/RSS.2023.XIX.026},
  url={https://www.roboticsproceedings.org/rss19/p026.html}
}

@inproceedings{rombach2022ldm,
  title={High-Resolution Image Synthesis with Latent Diffusion Models},
  author={Rombach, Robin and Blattmann, Andreas and Lorenz, Dominik and Esser, Patrick and Ommer, Bj{\"o}rn},
  booktitle={Proceedings of the IEEE/CVF Conference on Computer Vision and Pattern Recognition},
  year={2022},
  url={https://arxiv.org/abs/2112.10752}
}

@inproceedings{kang2025gflowvlm,
  title={Gflowvlm: Enhancing multi-step reasoning in vision-language models with generative flow networks},
  author={Kang, Haoqiang and Sachdeva, Enna and Gupta, Piyush and Bae, Sangjae and Lee, Kwonjoon},
  booktitle={2025 IEEE/CVF Conference on Computer Vision and Pattern Recognition (CVPR)},
  pages={3815--3825},
  year={2025},
  organization={IEEE}
}

@article{tu2026latent,
  title={Latent Reasoning with Normalizing Flows},
  author={Tu, Guancheng and Fu, Xiangjun and Yu, Suhao and Tang, Yao and Kang, Haoqiang and Qin, Lianhui and Zhang, Yizhe and Gu, Jiatao},
  journal={arXiv preprint arXiv:2606.06447},
  year={2026}
}

@article{kang2026ladi,
  title={LaDi-RL: Latent Diffusion Reasoning Prevents Entropy Collapse in Reinforcement Learning},
  author={Kang, Haoqiang and Zhang, Yizhe and Kuang, Nikki Lijing and Ma, Yi-An and Qin, Lianhui},
  journal={arXiv preprint arXiv:2602.01705},
  year={2026}
}

@article{kang2026scaffolding,
  title={Scaffolding Minds: Optimizing Latent Visual Target Representations for Multimodal Reasoning},
  author={Kang, Haoqiang and Chen, Yinpeng and Liu, Luyang and Andersen, Jesper Sparre and Ogale, Abhijit and Sun, Baochen and Hong, Lichan and Chi, Ed H},
  journal={arXiv preprint arXiv:2608.19669},
  year={2026}
}

@article{zhen2024threedvla,
  title={{3D-VLA}: A {3D} Vision-Language-Action Generative World Model},
  author={Zhen, Haoyu and Qiu, Xiaowen and Chen, Peihao and Yang, Jincheng and Yan, Xin and Du, Yilun and Hong, Yining and Gan, Chuang},
  journal={arXiv preprint arXiv:2403.09631},
  year={2024}
}

@article{duggal2026unite,
  title={End-to-End Training for Unified Tokenization and Latent Denoising},
  author={Duggal, Shivam and Bai, Xingjian and Wu, Zongze and Zhang, Richard and Shechtman, Eli and Torralba, Antonio and Isola, Phillip and Freeman, William T.},
  journal={arXiv preprint arXiv:2603.22283},
  year={2026},
  url={https://arxiv.org/abs/2603.22283}
}

@article{wei2025ovr,
  title={Open Vision Reasoner: Transferring Linguistic Cognitive Behavior for Visual Reasoning},
  author={Wei, Yana and Zhao, Liang and Sun, Jianjian and Lin, Kangheng and Yin, Jisheng and Hu, Jingcheng and Zhang, Yinmin and Yu, En and Lv, Haoran and Weng, Zejia and Wang, Jia and Han, Chunrui and Peng, Yuang and Han, Qi and Ge, Zheng and Zhang, Xiangyu and Jiang, Daxin and Patel, Vishal M.},
  journal={arXiv preprint arXiv:2507.05255},
  year={2025}
}

@article{li2023manipllm,
  title={ManipLLM: Embodied Multimodal Large Language Model for Object-Centric Robotic Manipulation},
  author={Li, Xiaoqi and Zhang, Mingxu and Geng, Yiran and Geng, Haoran and Long, Yuxing and Shen, Yan and Zhang, Renrui and Liu, Jiaming and Dong, Hao},
  journal={arXiv preprint arXiv:2312.16217},
  year={2023}
}

@inproceedings{chen2025ivtlr,
  title={Reasoning in the Dark: Interleaved Vision-Text Reasoning in Latent Space},
  author={Chen, Chao and Ma, Zhixin and Li, Yongqi and Hu, Yupeng and Wei, Yinwei and Li, Wenjie and Nie, Liqiang},
  booktitle={Findings of the Association for Computational Linguistics: ACL 2026},
  year={2026}
}

@article{chen2026lastr1,
  title={{LaST-R1}: Reinforcing Robotic Manipulation via Adaptive Physical Latent Reasoning},
  author={Chen, Hao and Liu, Jiaming and Yan, Zhonghao and Han, Nuowei and Zhang, Renrui and Gu, Chenyang and Gao, Jialin and Guo, Ziyu and Qian, Siyuan and Wang, Yinxi and Jia, Peng and Zhang, Shanghang and Heng, Pheng-Ann},
  journal={arXiv preprint arXiv:2604.28192},
  year={2026}
}

@inproceedings{yu2025flow,
  title={Flow of Reasoning: Training {LLMs} for Divergent Reasoning with Minimal Examples},
  author={Yu, Fangxu and Jiang, Lai and Kang, Haoqiang and Hao, Shibo and Qin, Lianhui},
  booktitle={Forty-second International Conference on Machine Learning},
  year={2025}
}

@article{bai2025qwen25vl,
  title={{Qwen2.5-VL} Technical Report},
  author={Bai, Shuai and Chen, Keqin and others},
  journal={arXiv preprint arXiv:2502.13923},
  year={2025}
}
\bibliographystyle{iclr2027_conference}

\clearpage
\appendix
\section{Method Details}
\label{app:method-details}

\subsection{Attention masks for parallel block training}
\label{app:attention-masks}

\begin{figure}[!htbp]
  \centering
  \includegraphics[width=\textwidth]{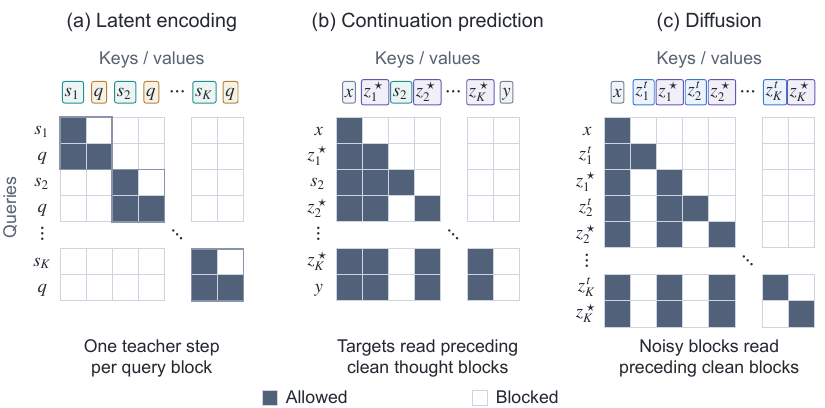}
  \caption{\textbf{Attention masks for parallel block training.} Rows attend to columns; each cell represents a token block. \textcolor{black}{Continuation prediction} interleaves clean thought blocks with target streams; diffusion interleaves noisy and clean thought blocks. Clean blocks never read target or noisy streams.}
  \label{fig:attention-masks}
\end{figure}

Figure~\ref{fig:attention-masks} illustrates the three masks for $K$ reasoning steps
packed into each pass; ellipses indicate omitted intermediate blocks. The restrictions apply at every layer,
preventing masked information from reaching predictions through other hidden states.

\paragraph{Latent encoding.}
The packed sequence contains pairs $[s_i;q]$ for all teacher steps.
Attention is isolated within each pair: the learnable block reads its own
teacher features and itself, but neither the task input nor other pairs.

\paragraph{\textcolor{black}{Continuation prediction}.}
The packed sequence is $[x;z_1^\star;s_2;z_2^\star;s_3;\ldots;z_K^\star;y]$,
where $s_i$ and $y$ label target prediction streams.
A prediction of $s_i^{\mathrm{GT}}$ can read $x$ and $z_{<i}^\star$;
the final-output prediction can read $x$ and all $z_{1:K}^\star$.
The conditioning stream preserves this prefix order and cannot read the
target streams. Different target streams are isolated from one another.
Within a text target, standard causal attention permits earlier tokens of
that target; continuous predictions cannot read their ground-truth values.

\paragraph{Diffusion.}
The packed sequence is $[x;z_1^t;z_1^\star;z_2^t;z_2^\star;\ldots;z_K^t;z_K^\star]$.
Each noisy block $z_i^t$ can read $x$, the clean prefix $z_{<i}^\star$,
and all tokens within itself. It cannot read its own clean target, later
clean blocks, or other noisy blocks. The clean conditioning stream follows
block-causal attention and cannot read noisy blocks. 

\section{\textcolor{black}{Experimental Details}}
\label{app:implementation}

\subsection{\textcolor{black}{Evaluation details}}
\label{app:protocols}

\paragraph{Benchmarks and scoring.}
All methods in Tables~\ref{tab:vlm-main} and~\ref{tab:backbone-transfer}
use the same evaluation examples within each benchmark.
Table~\ref{tab:evaluation-settings} lists the evaluation splits, sample
counts, and scoring units for the \rev{eleven} visual benchmarks~\citep{song2025visualpuzzles,masry2022chartqa,wu2024vstar,fu2024blink,tong2024mmvp,ray2024sat,tong2024cambrian,lu2024mathvista,wang2024mathvision,xu2025visulogic,hao2025emma}. VisualPuzzles and
VisuLogic use their released evaluation collections, even when the hosting
service names the split \texttt{train}. VisuLogic evaluation
uses the original 1,000-question benchmark~\citep{xu2025visulogic}, rather
than the current 1,003-row hosted collection. 

\paragraph{\newrev{Comparison protocol.}}
\newrev{On RLBench, we follow the LaST$_0$ protocol and take the baseline results from that paper~\citep{liu2026last0}. There, each baseline starts from its released checkpoint, is fine-tuned on the same 100 demonstrations per task, and is evaluated with the same rollouts; CoT-VLA is reimplemented on Janus-Pro. \method{} uses the same backbone, demonstrations, keyframe sampling, training schedule, and evaluation as LaST$_0$, so the two differ mainly in how latent reasoning is designed. On LIBERO, all methods use the standard training demonstrations and evaluation protocol of the benchmark. On the VLM side, every method in Table~\ref{tab:vlm-main} uses Qwen2.5-VL-7B and is scored on the same examples. Each baseline is trained with its own data and recipe, so Table~\ref{tab:vlm-main} compares complete methods rather than isolating the latent design. The analyses in Sections~\ref{sec:ablations} and~\ref{sec:mechanism-ablations} keep the backbone, training data, and teacher supervision fixed.}

\begin{table}[!htbp]
\centering
\caption{\textbf{Shared visual evaluation sets.} Splits, example counts, and scoring units used for the benchmark comparisons.}
\label{tab:evaluation-settings}
\small
\setlength{\tabcolsep}{4pt}
\renewcommand{\arraystretch}{1.10}
\begin{tabular*}{\textwidth}{@{\extracolsep{\fill}}llrl@{}}
\toprule
\textbf{Benchmark} & \textbf{Split} & \textbf{Size} & \textbf{Scoring unit} \\
\midrule
VisualPuzzles & Released collection & 1,168 & Question accuracy \\
ChartQA & Test & 2,500 & Relaxed answer accuracy \\
V$^*$ & Test & 191 & Question accuracy \\
BLINK-Jigsaw & Validation, Jigsaw & 150 & Question accuracy \\
MMVP & Full & 300 & Question accuracy \\
SAT & Real-image test & 150 & Question accuracy \\
CV-Bench & Test & 2,638 & Mean of 2D and 3D scores \\
MathVista & Testmini & 1,000 & Answer accuracy \\
MathVision & Test & 3,040 & Answer accuracy \\
VisuLogic & Original benchmark & 1,000 & Question accuracy \\
EMMA & Test, four subjects & 2,788 & Question accuracy \\
\bottomrule
\end{tabular*}
\end{table}

\subsection{\textcolor{black}{Training and inference details}}

\paragraph{VLM training.}
We train on Zebra-CoT~\citep{li2025zebracot}, using its interleaved text
and image steps as teacher supervision. We use AdamW with a peak learning rate of $10^{-5}$, cosine decay, a global batch
size of 64, and one epoch of training, informed by prior multimodal
fine-tuning recipes~\citep{li2025zebracot,hu2026colt}. We use bf16, a
maximum sequence length of 8,192, 3\% warmup, weight decay of 0.01, and
gradient clipping at 1.0. Each backbone retains its native image processor.
The unified encoder and shared reasoning backbone are optimized jointly;
teacher traces are omitted at evaluation.

\paragraph{Thought tokens and latent sampling.}
\label{app:latent-sampling}
We use four 512-dimensional thought tokens per block,
with learned projections to and from the backbone width, taking the token
budget from a latent-diffusion reference configuration~\citep{kang2025ladir}.
\newrev{The loss weights are $\lambda_y=2$ and $\lambda=5$.}
Training retains the annotated reasoning-step sequence\newrev{, and the number
of thought blocks is not fixed. At inference, the model decides when to stop:
after each block, it predicts the next token, and if this is the special token
\texttt{<BOT>}, it generates another block; otherwise, the latent chain ends.
For VLMs, the chain is followed by} greedy answer decoding with a 512-token limit.

For flow matching, we draw $\epsilon_i\sim\mathcal{N}(0,I)$ for each block and
$t\sim\mathcal{U}[0,1]$, set $z_i^t=(1-t)\epsilon_i+t\operatorname{sg}(z_i^\star)$, and use
$v_i^\star=z_i^\star-\epsilon_i$ as the velocity target~\citep{lipman2023flow}.
At inference, we integrate from $t=0$ to $1$ \rev{per block with one sampled
trajectory and no classifier-free guidance}.
\newrev{Table~\ref{tab:hyperparameters} collects these settings; VLM and VLA training use the same loss weights.}

\begin{table}[!htbp]
\centering
\caption{\textbf{Main hyperparameters.} Settings shared by both domains span the two columns.}
\label{tab:hyperparameters}
\footnotesize
\setlength{\tabcolsep}{4pt}
\renewcommand{\arraystretch}{1.10}
\begin{tabular*}{\textwidth}{@{\extracolsep{\fill}}lll@{}}
\toprule
\textbf{Setting} & \textbf{VLM} & \textbf{VLA} \\
\midrule
Backbone & Qwen2.5-VL-7B & MoT, $2\times$ Janus-Pro-1.5B (3.3B) \\
Training data & Zebra-CoT & RLBench 100 demos/task; LIBERO 500 demos/suite \\
Teacher modalities & Text, image & Future image, point cloud (RLBench), robot state \\
\midrule
Thought tokens per block & \multicolumn{2}{l}{4, each 512-dimensional} \\
Loss weights & \multicolumn{2}{l}{$\lambda_y=2$, $\lambda=5$} \\
Latent sampler & \multicolumn{2}{l}{Flow matching; one trajectory per block; no CFG} \\
Output decoding & Greedy, $\le$512 tokens & Flow-matching head, 10 steps, 7-D actions \\
Reasoning-to-action ratio & -- & 1:1, 1:2, 1:4 (train); 1:4 (eval) \\
\midrule
Optimizer & AdamW & AdamW \\
Peak learning rate & $10^{-5}$, cosine decay & $10^{-4}$, cosine decay to zero \\
Warmup / weight decay & 3\% / 0.01 & none / none \\
Global batch size & 64 & 64 \\
Training length & 1 epoch & 300 epochs \\
Precision / gradient clipping & \multicolumn{2}{l}{bf16 / 1.0} \\
Max.\ sequence length & 8,192 & -- \\
\bottomrule
\end{tabular*}
\end{table}

\paragraph{RLBench.}
We evaluate ten tasks with a Franka Panda in CoppeliaSim, using 100
waypoint-generated training demonstrations per task and keyframe
sampling~\citep{liu2026last0}. Inputs include a front-view RGB image resized
to $384\times384$, a point cloud sampled to 1,024 points, the task
instruction, and proprioception. We evaluate the final checkpoint with
20 rollouts per task and seed, giving 600 rollouts across three seeds.

\paragraph{LIBERO.}
We train a separate policy for each of the four suites, each containing
ten tasks and 500 training demonstrations. Inputs use third-person and
wrist RGB views resized to $384\times384$. We omit the point-cloud teacher
because this modality is unavailable, and sample intermediate teacher
steps every eight frames~\citep{liu2026last0}. We evaluate the final
checkpoint with 50 rollouts per task and seed, giving 6,000 rollouts
across the four suites and three seeds. Both benchmarks use their native
success conditions.

\paragraph{Seeds and aggregation.}
Our experiments use three random seeds. For question-level accuracy or
rollout success on $N$ examples per seed, the mean is
$\bar{s}=\frac{100}{3N}\sum_{r=1}^{3}\sum_{j=1}^{N}c_{rj}$,
where $c_{rj}\in\{0,1\}$ indicates correctness or success.
Grouped metrics retain the benchmark's aggregation rule: \rev{CV-Bench
averages its 2D and 3D scores.} For MMVP, we
report question-level accuracy over 300 questions, matching the
VLMEvalKit protocol used by the relevant comparison
methods~\citep{gu2025thinkmorph,li2026visualopsd}. Each answer is scored
independently against its correct option; the reported result is the mean
of the three seed-level accuracies.
Benchmark averages give equal weight to each reported benchmark or suite.
Relative gains are $100(s_{\rm ours}/s_{\rm baseline}-1)$; rounding follows
aggregation\rrev{, and gains are computed before table entries are rounded to one decimal}. RLBench Table~\ref{tab:rlbench} \rrev{reports the standard deviation across seeds in percentage points, following the baseline paper}~\citep{liu2026last0}. Reported gains describe differences
in mean scores.

\paragraph{VLA training.}
We train for 300 epochs with AdamW, using a peak learning rate of $10^{-4}$,
cosine decay to zero, no weight decay or warmup, and gradient clipping at
1.0~\citep{liu2026last0}. Training uses eight NVIDIA A800 GPUs with bf16
and DeepSpeed ZeRO-1. The batch size is eight per device and 64 globally,
without gradient accumulation. Actions are seven-dimensional, with ten
flow-matching steps in the action head.
We mix reasoning-to-action ratios of 1:1, 1:2, and 1:4 during training and
use 1:4 at evaluation. The latent diffusion objective is defined in
Section~\ref{sec:method}.

\paragraph{Inference speed.}
All inference rates in Table~\ref{tab:rlbench}, including those of the
baselines, are measured on the same single NVIDIA RTX 4090, without action
chunking, using one reasoning update per four action decisions~\citep{liu2026last0}.
We use batch size one and bf16, with
100 warmup decisions followed by 1,000 timed decisions per seed.
Wall-clock timing is synchronized with CUDA before and after the timed
sequence and includes observation preprocessing, host-to-device transfer,
reasoning updates, and action prediction; simulator stepping and rendering
are excluded. Each method uses the same inputs and timing boundary.
Throughput is the number of decisions divided by elapsed time, averaged
across three seeds. The reported 16.5 Hz corresponds to 60.6 ms per decision.

\subsection{Ablation protocols}
\label{app:ablation-protocols}

\paragraph{Latent target supervision and encoding paths.}
The extended design in Figure~\ref{fig:ablation-objective} crosses two intermediate objectives with whether the LLM encoder is shared across modalities. In the \emph{separate} path, each modality is encoded by its own LLM encoder, which has the same architecture, learnable queries, and output projection as the unified encoder. Recon and Cont train these encoders jointly with the reasoner. The Frozen (separate) control keeps each modality's encoder frozen, so the latent targets stay fixed, and trains only the downstream reasoner. The Separate baseline in Figure~\ref{fig:ablation-sharing} is this same continuation-trained separate variant; all shared scores use identical source values. The \emph{unified} path uses the shared queries
and LLM encoding pass described in Section~\ref{sec:construct}.
All five variants use the same number and width of thought tokens.
The learned variants share data, training steps, diffusion loss, and
$\lambda_y\mathcal L_{\mathrm{out}}$. Reconstruction predicts source teacher
features, whereas continuation predicts later teacher steps; text CE and
continuous L2 use the same reductions and weights across paths. Target features
remain fixed. Because the separate path uses one encoder per modality, it has at least as many encoder parameters as the unified path; the two paths differ only in whether the encoder is shared across modalities.

\paragraph{Aggregation and rounding.}
Figure~\ref{fig:ablation-objective} uses the same three-seed aggregation as the
main tables: $N=191$, $300$, $1{,}000$, and $2{,}788$ questions per seed for
V$^*$, MMVP, MathVista, and EMMA; $N=2{,}000$ and $200$ rollouts per seed for
LIBERO and RLBench. Scores are rounded to one decimal after aggregation;
cross-benchmark differences use unrounded scores and weight benchmarks equally.

\paragraph{Thought-token prediction.}
\rrev{In Figure~\ref{fig:ablation-diffusion}, all three objectives are trained jointly
with the unified encoder, using the same continuation loss and loss weights; only
the thought-token prediction objective changes.} L2 and cosine regression directly predict the
next clean thought block. Flow matching predicts the velocity along the
noisy-to-clean path, using the sampler specified in Appendix~\ref{app:protocols}.
Each method conditions on the same input and preceding thought blocks and
supplies its predicted tokens to the downstream reasoner. This comparison
reports downstream performance for the respective prediction procedures.

\subsection{\textcolor{black}{Additional experiments}}
\label{app:additional-experiments}

\paragraph{Reasoning across backbones.}
We evaluate \method{} with BAGEL-7B, Qwen3-VL-8B, and Janus-Pro at
1.5B and 7B to test whether the gains extend beyond the main backbone
(Table~\ref{tab:backbone-transfer}). On the \rev{seven} vision-centric benchmarks,
the average improves over the corresponding base models by
\rev{29.5\%}, \rev{12.4\%}, \rev{35.9\%}, and \rev{51.5\%},
respectively. With BAGEL, \method{} also improves the average over
Visual-OPSD by 2.5\%, leading on \rev{six of seven} benchmarks;
with Qwen3-VL, it improves over CoLT by \rev{13.3\%}.
On BAGEL's mathematical and logical benchmarks, the average gains are
27.9\% over the base model and 20.0\% over Uni-CoT~\citep{qin2025unicot},
including a 28.3\% gain on EMMA over Uni-CoT.
These improvements across model families and sizes suggest that shared
latent reasoning is useful beyond a single backbone.

\begin{table}[!htbp]
\centering
\caption{\textbf{Reasoning across backbones.} Accuracy (\%) on \textbf{(a)} \rev{seven} vision-centric benchmarks and \textbf{(b)} four mathematical and logical benchmarks. Bold marks the best result within each backbone.}
\label{tab:backbone-transfer}
\small
\setlength{\tabcolsep}{2.0pt}
\renewcommand{\arraystretch}{1.10}
\begin{tabular*}{\textwidth}{@{\extracolsep{\fill}}l*{7}{r}!{\tablesummarysep}>{\color{optiontitleink}}r@{}}
\toprule
\multicolumn{9}{@{}l}{\tablegroup{(a) Vision-centric reasoning}} \\
\midrule
\textbf{\textcolor{optiontitleink}{Method}} & \textbf{VisPuz.} & \textbf{ChartQA} & \textbf{V$^*$} & \textbf{BLINK-J} & \textbf{MMVP} & \textbf{SAT} & \textbf{CV-B} & \textbf{Avg.} \\
\midrule
\rowcolor{tableblue}\multicolumn{9}{l}{\tablegroup{BAGEL-7B}} \\
Base & \bluecell{35.0} & \bluecell{61.8} & \bluecell{55.5} & 67.3 & 70.3 & 44.7 & 76.0 & \rev{58.7} \\
Visual-OPSD & \bluecell{\textbf{86.0}} & 78.8 & 64.9 & 77.3 & 77.3 & 54.0 & 80.6 & \rev{74.1} \\
ThinkMorph & 79.0 & \bluecell{78.1} & 67.0 & 72.0 & 80.3 & 52.7 & 80.8 & \rev{72.8} \\
\textbf{Uni-LaDiR (Ours)} & 75.1 & \textbf{83.8} & \textbf{69.1} & \textbf{80.0} & \textbf{83.0} & \textbf{54.7} & \textbf{86.4} & \rev{\textbf{76.0}} \\
\midrule
\rowcolor{tableblue}\multicolumn{9}{l}{\tablegroup{Qwen3-VL-8B}} \\
Base & \bluecell{37.0} & 82.6 & 84.3 & 68.7 & 77.7 & 54.0 & 85.6 & \rev{70.0} \\
CoLT & 39.0 & \bluecell{74.7} & 82.2 & 70.0 & 78.0 & 58.0 & 84.0 & \rev{69.4} \\
\textbf{Uni-LaDiR (Ours)} & \textbf{60.3} & \textbf{86.1} & \textbf{85.9} & \textbf{78.0} & \textbf{84.0} & \textbf{67.3} & \textbf{89.1} & \rev{\textbf{78.7}} \\
\midrule
\rowcolor{tableblue}\multicolumn{9}{l}{\tablegroup{Janus-Pro-1.5B}} \\
Base & 28.0 & 52.0 & 41.9 & 48.0 & 58.0 & 24.0 & 64.0 & \rev{45.1} \\
\textbf{Uni-LaDiR (Ours)} & \textbf{58.7} & \textbf{68.2} & \textbf{57.1} & \textbf{64.0} & \textbf{69.3} & \textbf{38.7} & \textbf{73.1} & \rev{\textbf{61.3}} \\
\midrule
\rowcolor{tableblue}\multicolumn{9}{l}{\tablegroup{Janus-Pro-7B}} \\
Base & 33.5 & 43.1 & 38.2 & 50.7 & 63.3 & 22.0 & 67.8 & \rev{45.5} \\
\textbf{Uni-LaDiR (Ours)} & \textbf{69.8} & \textbf{75.2} & \textbf{64.4} & \textbf{70.0} & \textbf{76.3} & \textbf{46.7} & \textbf{80.3} & \rev{\textbf{68.9}} \\
\bottomrule
\end{tabular*}
\par\smallskip
\begin{tabular*}{\textwidth}{@{\extracolsep{\fill}}l*{4}{r}!{\tablesummarysep}>{\color{optiontitleink}}r@{}}
\toprule
\multicolumn{6}{@{}l}{\tablegroup{(b) Mathematical and logical reasoning: BAGEL-7B}} \\
\midrule
\textbf{\textcolor{optiontitleink}{Method}} & \textbf{MathVista} & \textbf{MathVision} & \textbf{VisuLogic} & \textbf{EMMA} & \textbf{Avg.} \\
\midrule
Base & 72.5 & 36.0 & 28.9 & 28.7 & 41.5 \\
Uni-CoT & 73.3 & 48.5 & 25.0 & 30.2 & 44.2 \\
Bagel-Zebra-CoT & 72.1 & 43.5 & 0.0 & 20.6 & 34.0 \\
ThinkMorph & 67.8 & 41.5 & 6.5 & 22.4 & 34.6 \\
\textbf{Uni-LaDiR (Ours)} & \textbf{79.8} & \textbf{57.9} & \textbf{35.9} & \textbf{38.7} & \textbf{53.1} \\
\bottomrule
\end{tabular*}
\end{table}

\begin{figure}[!htbp]
  \centering

  \includegraphics[width=\textwidth]{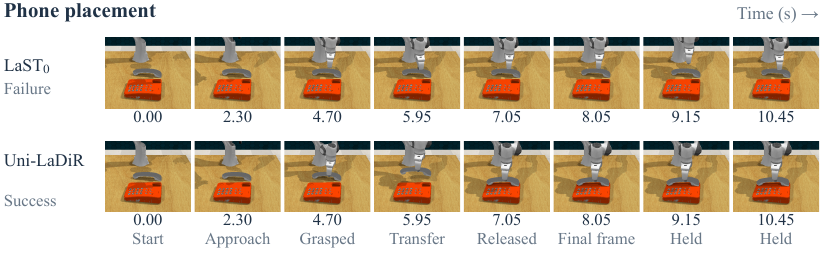}
  \caption{\textbf{Phone placement in RLBench.} LaST$_0$ does not grasp the phone within ten recorded decisions (top), while \method{} completes placement (bottom). Columns share simulation timestamps, excluding inference latency; the final frame is held after each recording ends.}
  \label{fig:rlbench-qualitative}
\end{figure}

\begingroup
\color{black}
\paragraph{Latent interventions.}
Prior work shows that latent reasoning methods can bypass intermediate latents without losing accuracy~\citep{cui2026latentsupervision,viveiros2026holding}. We therefore test whether \method{} uses its generated thought tokens. We zero the latent chain, shuffle its blocks, or replace it with a chain from another example, while keeping the input and sequence length unchanged:
\begin{equation}
\widetilde z_{1:K}=\begin{cases}
z_{1:K}, & \text{intact},\\
0, & \text{zeroed},\\
[z_{\pi(1)};\ldots;z_{\pi(K)}], & \text{shuffled},\\
z_{1:K}^{(j)}, & \text{cross-example}.
\end{cases}
\label{eq:latent-interventions}
\end{equation}
Here $\pi$ permutes the latent blocks while preserving their internal token order, and $z_{1:K}^{(j)}$ comes from another example with the same latent shape. We recompute the output with fresh hidden states and pair decoding or rollout seeds across conditions. For VLA evaluation, we apply the intervention at every reasoning update under the same initial scene.

\begin{table}[!htbp]
\centering
\color{black}
\caption{\textcolor{black}{\textbf{Latent-intervention results.} We modify the generated latent chain while keeping the input and sequence length fixed. $\Delta$ denotes the change from the intact chain in percentage points.}}
\label{tab:latent-interventions}
\begin{tabular}{lrrrr}
\toprule
& \multicolumn{2}{c}{V$^*$} & \multicolumn{2}{c}{RLBench} \\
\cmidrule(lr){2-3}\cmidrule(lr){4-5}
Latent chain & Accuracy (\%) & $\Delta$ & Success (\%) & $\Delta$ \\
\midrule
Intact generated & \rrev{\textbf{89.5}} & --- & \textbf{87.0} & --- \\
Zeroed & \rrev{55.8} & \rrev{$-33.7$} & 55.0 & $-32.0$ \\
Shuffled blocks & \rrev{65.8} & \rrev{$-23.7$} & 65.0 & $-22.0$ \\
Cross-example & \rrev{50.8} & \rrev{$-38.7$} & 51.0 & $-36.0$ \\
\bottomrule
\end{tabular}
\end{table}

Zeroing the latent chain reduces performance by \rrev{32.0--33.7} percentage points, while shuffling its blocks causes a \rrev{22.0--23.7} point drop (Table~\ref{tab:latent-interventions}). Replacing the chain with one from another example has the largest effect, decreasing performance by \rrev{36.0--38.7} points. Together with the supervision ablation in Figure~\ref{fig:ablation-objective}, these results show that continuation prediction learns useful latent representations: the generated thought tokens retain input-specific, order-sensitive information used for prediction.
\endgroup
\paragraph{Qualitative manipulation comparison.}
Figure~\ref{fig:rlbench-qualitative} compares a LaST$_0$ rollout with a
successful \method{} rollout from the same initial scene. LaST$_0$
approaches the phone but does not secure a grasp within ten decisions.
\method{} grasps the phone, transfers it to the base, and releases
it, reaching success at $7.05$ seconds of simulation time. This selected
episode illustrates a difference at the grasping stage; aggregate success
rates are reported in Table~\ref{tab:rlbench}.

\newrev{
\section{Limitations}
\label{app:limitations}
Our experiments focus on whether reasoning in a shared latent space helps, and we leave a closer study of the teacher CoTs themselves to future work. Beyond which modalities are shared (Section~\ref{sec:ablations}), we do not analyze how the teacher CoTs affect the learned reasoning, for example the granularity of teacher steps or the source and quality of the traces. In our setting, the VLA teachers are future images, point clouds, and robot states taken from the demonstrations, and the VLM teachers are the interleaved traces in Zebra-CoT; other sources, such as tactile signals or model-generated traces, may shape the latent space differently.
}

\end{document}